\documentclass[letterpaper,10pt,journal,twoside]{IEEEtran}  

\usepackage{multicol}
\usepackage[colorlinks=true, allcolors=blue]{hyperref}

\usepackage{amsmath,amsfonts,amssymb,amsthm} 
\usepackage{graphicx}
\usepackage{xspace}
\usepackage{multirow}

\usepackage{graphicx}
\usepackage{xr}
\graphicspath{{./}}

\usepackage{comment}
\usepackage{siunitx}
\usepackage{relsize}
\usepackage{ifthen}
\usepackage[colorinlistoftodos]{todonotes}

\usepackage[caption=false]{subfig}

\usepackage[vlined,ruled,linesnumbered]{algorithm2e}
\usepackage{graphics} 
\usepackage{rotating}
\usepackage{color}
\usepackage{enumerate}
\usepackage[T1]{fontenc}
\usepackage{psfrag}
\usepackage{epsfig} 
\usepackage{booktabs}
\usepackage{graphicx,url}
\usepackage{multirow}
\usepackage{array}
\usepackage{latexsym}
\usepackage{amsfonts}
\usepackage{amsmath}
\usepackage{amssymb}
\usepackage{xstring}
\usepackage[noend]{algorithmic}
\usepackage{multirow}
\usepackage{xcolor}
\usepackage{prettyref}
\usepackage{flexisym}
\usepackage{bigdelim}
\usepackage{breqn} 
\usepackage{listings}
\let\labelindent\relax
\usepackage{enumitem}
\usepackage{xspace}
\usepackage{bm}
\graphicspath{{./figures/}}
\usepackage{tikz}
\usetikzlibrary{matrix,calc}

\usepackage{mdwlist}

\let\itemize\undefined
\makecompactlist{itemize}{stditemize}

\newrefformat{prob}{Problem\,\ref{#1}}
\newrefformat{def}{Definition\,\ref{#1}}
\newrefformat{sec}{Section\,\ref{#1}}
\newrefformat{sub}{Section\,\ref{#1}}
\newrefformat{prop}{Proposition\,\ref{#1}}
\newrefformat{app}{Appendix\,\ref{#1}}
\newrefformat{alg}{Algorithm\,\ref{#1}}
\newrefformat{cor}{Corollary\,\ref{#1}}
\newrefformat{thm}{Theorem\,\ref{#1}}
\newrefformat{lem}{Lemma\,\ref{#1}}
\newrefformat{fig}{Fig.\,\ref{#1}}
\newrefformat{tab}{Table\,\ref{#1}}

\newtheorem{theorem}{Theorem}

\newtheorem{lemma}[theorem]{Lemma}

\newtheorem{proposition}[theorem]{Proposition}
\newtheorem{remark}[theorem]{Remark}
\newtheorem{example}[theorem]{Example}

\newcommand{\bdmath}{\begin{dmath}}
\newcommand{\edmath}{\end{dmath}}
\newcommand{\beq}{\begin{equation}}
\newcommand{\eeq}{\end{equation}}
\newcommand{\bdm}{\begin{displaymath}}
\newcommand{\edm}{\end{displaymath}}
\newcommand{\bea}{\begin{eqnarray}}
\newcommand{\eea}{\end{eqnarray}}
\newcommand{\beal}{\beq \begin{array}{ll}}
\newcommand{\eeal}{\end{array} \eeq}
\newcommand{\beas}{\begin{eqnarray*}}
\newcommand{\eeas}{\end{eqnarray*}}
\newcommand{\ba}{\begin{array}}
\newcommand{\ea}{\end{array}}
\newcommand{\bit}{\begin{itemize}}
\newcommand{\eit}{\end{itemize}}
\newcommand{\ben}{\begin{enumerate}}
\newcommand{\een}{\end{enumerate}}

\newcommand{\calR}{{\cal R}}

\newcommand{\calX}{{\cal X}}

\newcommand{\etal}{\emph{et~al.}\xspace}

\newcommand{\M}[1]{{\bm #1}} 
\renewcommand{\boldsymbol}[1]{{\bm #1}}

\newcommand{\hide}[1]{}
\newcommand{\wrt}{w.r.t.\xspace}

\newcommand{\hiddenText}{{\color{gray} hidden text.}}
\newcommand{\hideWithText}[1]{\hiddenText}

\newcommand{\kron}{\otimes}

\DeclareMathOperator*{\argmin}{arg\,min}

\newcommand{\tran}{^{\mathsf{T}}}

\newcommand{\trace}[1]{\mathrm{tr}\left(#1\right)}

\newcommand{\Real}[1]{ { {\mathbb R}^{#1} } }

\newcommand{\at}[1]{^{(#1)}}

\newcommand{\SOthree}{\ensuremath{\mathrm{SO}(3)}\xspace}

\newcommand{\MA}{\M{A}}
\newcommand{\MB}{\M{B}}

\newcommand{\MP}{\M{P}}
\newcommand{\MQ}{\M{Q}}

\newcommand{\MR}{\M{R}}

\newcommand{\MF}{\M{F}}

\newcommand{\va}{\boldsymbol{a}}

\newcommand{\vf}{\boldsymbol{f}}
\newcommand{\vg}{\boldsymbol{g}}

\newcommand{\vq}{\boldsymbol{q}}
\newcommand{\vr}{\boldsymbol{r}}

\newcommand{\vv}{\boldsymbol{v}}
\newcommand{\vt}{\boldsymbol{t}}
\newcommand{\vxx}{\boldsymbol{x}} 
\newcommand{\vy}{\boldsymbol{y}}
\newcommand{\vw}{\boldsymbol{w}}

\newcommand{\scenario}[1]{{\smaller \sf#1}\xspace}

\newcommand{\gtwoo}{{\smaller\sf g2o}\xspace}

\newcommand{\intel}{\scenario{INTEL}}

\newcommand{\csail}{\scenario{CSAIL}}

\newcommand{\blue}[1]{{\color{blue}#1}}

\newcommand{\linkToPdf}[1]{\href{#1}{\blue{(pdf)}}}
\newcommand{\linkToPpt}[1]{\href{#1}{\blue{(ppt)}}}
\newcommand{\linkToCode}[1]{\href{#1}{\blue{(code)}}}
\newcommand{\linkToWeb}[1]{\href{#1}{\blue{(web)}}}
\newcommand{\linkToVideo}[1]{\href{#1}{\blue{(video)}}}
\newcommand{\linkToMedia}[1]{\href{#1}{\blue{(media)}}}
\newcommand{\award}[1]{\xspace} 

\newcommand{\vz}{\boldsymbol{z}}

\newcommand{\ransac}{\scenario{RANSAC}}
\newcommand{\GNC}{\scenario{GNC}}
\newcommand{\GNCGM}{\scenario{GNC-GM}}
\newcommand{\GNCTLS}{\scenario{GNC-TLS}}
\newcommand{\sesync}{\scenario{SE-Sync}}
\newcommand{\PCR}{\scenario{P-REG}}
\newcommand{\GReg}{\scenario{G-REG}}
\newcommand{\PGO}{\scenario{PGO}}
\newcommand{\shape}{\scenario{SA}}
\newcommand{\adapt}{\scenario{ADAPT}}
\newcommand{\dcs}{\scenario{DCS}}
\newcommand{\pcm}{\scenario{PCM}}
\newcommand{\bunny}{\scenario{Bunny}}
\newcommand{\PASCALplus}{\scenario{PASCAL+}}
\newcommand{\FGCar}{\scenario{FG3DCar}}
\newcommand{\TEASER}{\scenario{TEASER}}

\newcommand{\gPoly}{\scenario{GloptiPoly 3}}

\newcommand{\sumAllPointsi}{\sum_{i=1}^{N}}

\newcommand{\barc}{\bar{c}}
\newcommand{\barcsq}{\barc^2}
\newcommand{\vectorize}[1]{\mathrm{vec}\left(#1\right)}

\newcommand{\bmat}{\left[ \begin{array}}
\newcommand{\emat}{\end{array} \right]}

\newcommand{\barvz}{\bar{\vz}}
\newcommand{\barMB}{\bar{\MB}}

\newcommand{\ie}{{i.e.},\xspace}
\newcommand{\eg}{{e.g.},\xspace}

\newcommand{\tldvt}{\tilde{\vt}}
\newcommand{\tldMR}{\tilde{\MR}}
\newcommand{\tldvz}{\tilde{\vz}}
\newcommand{\tldMB}{\tilde{\MB}}

\newcommand{\bal}{\begin{align}}
\newcommand{\eal}{\end{align}}

\newcommand{\hatr}{\hat{r}}

\newcommand{\revise}[1]{#1}
\newcommand{\finalize}[1]{#1}

\newcommand{\myParagraph}[1]{{\bf #1.}}

\newcommand{\isExtended}[2]{#2} 

\title{\huge{Graduated Non-Convexity for Robust Spatial Perception: \\ From Non-Minimal Solvers to Global Outlier Rejection}}

\author{Heng Yang, Pasquale Antonante, Vasileios Tzoumas, Luca Carlone

\thanks{This work was partially funded by ARL DCIST CRA W911NF-17-2-0181, ONR RAIDER N00014-18-1-2828, and Lincoln Laboratory's Resilient Perception in Degraded Environments program.}

\thanks{
The authors are with the Laboratory for 
Information \& Decision Systems (LIDS), Massachusetts Institute of Technology, Cambridge, USA, 
{\tt\scriptsize \{hankyang,antonap,vtzoumas,lcarlone\}@mit.edu}}
  \vspace{-1cm}
}
 
\begin{document}

\maketitle
\begin{tikzpicture}[overlay, remember picture]
\path (current page.north east) ++(-4.2,-0.2) node[below left] { 
This paper has been accepted for publication in the IEEE Robotics and Automation Letters.
};
\end{tikzpicture}
\begin{tikzpicture}[overlay, remember picture]
\path (current page.north east) ++(-5.3,-0.6) node[below left] {
Please cite the paper as: H. Yang, P. Antonante, V. Tzoumas, and L. Carlone,
};
\end{tikzpicture}
\begin{tikzpicture}[overlay, remember picture]
\path (current page.north east) ++(-2.3,-1) node[below left] {
``Graduated Non-Convexity for Robust Spatial Perception: From Non-Minimal Solvers to Global Outlier Rejection'',
};
\end{tikzpicture}
\begin{tikzpicture}[overlay, remember picture]
\path (current page.north east) ++(-7.3,-1.4) node[below left] {
 \emph{IEEE Robotics and Automation Letters (RA-L)}, 2020.
};
\end{tikzpicture}

\vspace{-3.7mm}
\begin{abstract}
Semidefinite Programming (SDP) and Sums-of-Squ- ares (SOS) relaxations have 
led to
\emph{certifiably} optimal \emph{non-minimal} solvers for several robotics and computer vision problems. 
However, most non-minimal solvers rely on least squares formulations, and, as a result, are 
brittle against
outliers.
%
While a standard approach to regain robustness against outliers 
is to use robust cost functions, the latter 
typically introduce 
other non-convexities, 
preventing 
the use of 
existing 
non-minimal solvers.
In this paper, we 
enable the simultaneous use of 
non-minimal solvers and robust estimation
by providing a general-purpose approach for robust \emph{global} estimation, which can be applied to any {problem where a non-minimal solver is available for the outlier-free case.}
To this end,
we leverage the Black-Rangarajan duality between robust estimation and outlier processes (which has been traditionally applied to early vision problems), and show that
\emph{graduated non-convexity} (\GNC) 
can be used 
in conjunction with non-minimal solvers 
to compute robust solutions, without requiring 
an initial guess. 
\revise{Although \GNC's global optimality cannot be guaranteed, }we demonstrate\revise{~the empirical robustness of} the resulting \emph{robust non-minimal solvers} in applications, including 
point cloud and mesh registration, pose graph optimization, and image-based object pose estimation (also called \emph{shape alignment}).
Our solvers are robust to {70-80\%} of outliers, 
outperform \ransac, 
are more accurate than specialized \emph{local} solvers, and 
faster than specialized \emph{global} solvers.
\finalize{We also propose the \emph{first} certifiably optimal non-minimal solver for shape alignment using SOS relaxation.}
 %
\end{abstract}


\section{Introduction}
\label{sec:introduction}
\IEEEPARstart{R}{obust}
estimation is a crucial tool for robotics and computer vision, 
being
concerned with the estimation of unknown quantities 
(\eg the state of a robot, or of variables describing the external world) 
from noisy and potentially corrupted measurements. Corrupted measurements (\ie \emph{outliers}) can be caused by sensor malfunction, but are  more commonly associated with 
incorrect data association \mbox{and model misspecification~\cite{Cadena16tro-SLAMsurvey,Yang19rss-teaser}.} 

In the outlier-free case, common estimation problems 
are formulated as a least squares optimization: 
\bea \label{eq:leastSquares}
\min_{\vxx \in \calX} \;\sumAllPointsi r^2(\vy_i,\vxx),
\eea 
where $\vxx$ is the variable we want to estimate (\eg the pose of an unknown object); 
$\calX$ is the domain of $\vxx$ (\eg the set of 3D poses);
$\vy_i$ ($i=1,\ldots,N$) are given measurements (\eg pixel observations of points belonging to the object); 
  and the function $r(\vy_i,\vxx)$ is the \emph{residual} error for the $i$-th measurement,
quantifying the mismatch between the expected measurement at an estimate $\vxx$ and the actual measurement~$\vy_i$.
In typical robotics and computer vision applications, 
\revise{~the least squares optimization}~\eqref{eq:leastSquares} is difficult to solve globally, due to the nonlinearity of the 
residual errors and the nonconvexity of the domain~$\calX$. Despite these challenges, 
the research community has developed closed-form solutions and\revise{~globally optimal} solvers for many such problems.
Specifically, 
while closed-form solutions are rare~\cite{Horn87josa}, 
\emph{Semidefinite Programming} (SDP) and \emph{Sums-of-Squares} (SOS)~\cite{Blekherman12Book-sdpandConvexAlgebraicGeometry}
relaxations have been recently shown to 
  be a powerful tool to obtain \emph{certifiably optimal} solutions to relevant instances of problem~\eqref{eq:leastSquares}, 
  ranging from pose graph optimization~\cite{Carlone16tro-duality2D,Rosen18ijrr-sesync},
   rotation averaging~\cite{Eriksson18cvpr-strongDuality},
   anisotropic registration~\cite{Briales17cvpr-registration}, 
   two-view geometry~\cite{Briales18cvpr-global2view}, 
   and PnP~\cite{Agostinho2019arXiv-cvxpnpl}.
   The resulting techniques are commonly referred to as \emph{non-minimal solvers}, to contrast them against 
   methods that solve problem~\eqref{eq:leastSquares} using only a small (minimal) subset of measurements $\vy_i$ 
   (see related work in Section~\ref{sec:relatedWork}).

Unfortunately, in the presence of outliers, problem~\eqref{eq:leastSquares}'s solution provides a 
poor estimate for 
$\vxx$. This limits the applicability of existing non-minimal solvers, 
 since they can be applied only after the outliers have been removed. 
 Yet,
 the theory of robust estimation  
 suggests regaining robustness by substituting the quadratic cost in\revise{~the least squares problem}~\eqref{eq:leastSquares} 
 with a \emph{robust cost} $\rho(\cdot)$:
\bea \label{eq:generalEstimation}
\min_{\vxx \in \calX} \; \sumAllPointsi \rho( r(\vy_i,\vxx) ).
\eea 
%
For instance, $\rho(\cdot)$ can be a Huber loss, a truncated least squares cost, or a Geman-McClure cost~\cite{Black96ijcv-unification}.
To date, the application of non-minimal solvers  to eq.~\eqref{eq:generalEstimation} has been limited. 
Some of the non-minimal solvers designed for~\eqref{eq:leastSquares} can be extended to include a \emph{convex} robust cost function, such as the Huber loss~\cite{Carlone18ral-robustPGO2D}; however, it is known that convex losses have a low breakdown point and are still sensitive to gross outliers~\cite{Lajoie19ral-DCGM}. 
In rare cases, the literature provides \emph{robust non-minimal solvers} that achieve 
global solutions to specific instances of problem~\eqref{eq:generalEstimation}, 
such as robust registration~\cite{Yang19rss-teaser,Yang19iccv-QUASAR}. 
However, 
these solvers cannot be easily extended to other estimation problems, and they rely on solving large SDPs, which is currently
  impractical for large-scale problems.

{\bf Contributions.} In this paper, we aim to reconcile non-minimal solvers and robust estimation, by providing a 
\textit{general-purpose}\footnote{For a given problem~\eqref{eq:generalEstimation}, we only assume the existence of a non-minimal solver for the 
corresponding outlier-free problem~\eqref{eq:leastSquares}.} algorithm to solve problem~\eqref{eq:generalEstimation} without requiring an initial guess. 
We achieve this by combining  non-minimal solvers with an approach known as \emph{graduated non-convexity} (\GNC). 
In contrast, standard algorithms for problem~\eqref{eq:generalEstimation} rely on iterative optimization to refine a given initial guess, which causes the result to be brittle when the quality of the  guess is poor. 
Particularly, we propose three contributions.

First, we revisit the Black-Rangarajan duality~\cite{Black96ijcv-unification} between robust estimation and outlier processes.
  We also revisit the use of \emph{graduated non-convexity} (\GNC) as a general tool to solve a non-convex 
  optimization problem without an initial guess.
  While Black-Rangarajan duality and \GNC have been used in early vision problems, such as stereo reconstruction, 
  image restoration and segmentation, and optical flow, 
 we show that combining them with non-minimal solvers allows solving \emph{spatial perception} problems, ranging from 
 mesh registration, pose graph optimization, and image-based object pose estimation.

Our second contribution is to
tailor Black-Rangarajan duality and \GNC to the 
 \textit{Geman-McClure} and \textit{truncated least squares} costs.
We show how to optimize these functions  by alternating two steps: 
a \emph{variable update}, which solves a \emph{weighted} least squares problem using non-minimal solvers; 
 and a \emph{weight update}, \mbox{which updates the outlier process in closed~form.} 

Our approach requires 
a non-minimal solver, but currently there is no such solver for 
image-based object pose estimation (also known as \emph{shape alignment}).
Our third contribution is to present a novel non-minimal solver for shape alignment.
While related techniques propose approximate relaxations~\cite{Zhou17pami-shapeEstimationConvex}, we 
provide a 
\emph{certifiably optimal} solution 
using SOS relaxation. 

We demonstrate our \emph{robust non-minimal solvers} on point cloud registration (\PCR), mesh registration (as known as \emph{generalized} registration, \GReg), pose graph optimization (\PGO), and shape alignment (\shape).
Our solvers  are robust to {70-80\%} outliers, 
outperform \ransac, 
are more accurate than specialized \emph{local} solvers, and 
faster than \emph{global} solvers.

\section{Related Work}
\label{sec:relatedWork}

\subsection{Outlier-free Estimation}
\label{sec:rw_outlierFree}
{\bf Minimal Solvers.} Minimal solvers use the smallest number of measurements necessary to estimate $\vxx$.
Examples include the 2-point solver for the Wahba problem (a rotation-only variant of \PCR)~\cite{
Markley14book-fundamentalsAttitudeDetermine}, the 3-point solver for \PCR~\cite{Horn87josa}, and the 12-point solver~\cite{Khoshelham16jprs-ClosedformSolutionPlaneCorrespondences} for \GReg with point-to-plane correspondences.
Notably, the approach~\cite{Kukelova13thesis-algebraicTechniquesinCV} enables minimal solvers for a growing number of estimation problems. 
\isExtended{For a more exhaustive list of minimal solvers, we refer the reader to~\cite{PajdlaXXwebsite-minimalProblemsInVision}.}{}

{\bf Non-minimal Solvers.} Minimal solvers do not leverage data redundancy, and, consequently, can be sensitive to measurement noise. Therefore, non-minimal solvers have been developed. Typically, 
non-minimal solvers 
assume Gaussian measurement noise, which results in a {least squares} optimization framework. 
In some cases, the resulting optimization problems can be solved in closed form,  
\eg Horn's method~\cite{Horn87josa} for~\PCR, or
 PLICP~\cite{Censi08icra}~for~\GReg.
Generally, however, the resulting optimization problems are hard\revise{~and only locally optimal solutions can be obtained~\cite{Kuemmerle11icra}}. For global optimization, researchers have developed exponential-time methods, such as \emph{Branch and Bound} (BnB). 
Hartley and Kahl~\cite{Hartley09ijcv-globalRotationRegistration} introduce a BnB search over the rotation space to globally solve several vision problems. Olsson \etal~\cite{Olsson09pami-bnbRegistration} develop optimal solutions for \GReg.
Recently, \emph{Semidefinite Programing} (SDP) and \emph{Sums of Squares} (SOS) relaxations~\cite{Blekherman12Book-sdpandConvexAlgebraicGeometry}
have been used to develop
polynomial-time 
algorithms with certifiable optimality guarantees. 
Briales and Gonzalez-Jimenez~\cite{Briales17cvpr-registration}
solve~\GReg using SDP.
Carlone~\etal~\cite{Carlone16tro-duality2D,Carlone15iros-duality3D} use Lagrangian duality and SDP relaxations for \PGO.
 Rosen~\etal~\cite{Rosen18ijrr-sesync} develop \sesync, a fast solver for the relaxation in~\cite{Carlone16tro-duality2D,Carlone15iros-duality3D}.
 Mangelson~\etal~\cite{Mangelson19icra-PGOwithSBSOS} apply the sparse bounded-degree variant of 
 \isExtended{SOS relaxation~\cite{Weisser18mpc-SBSOS},}{SOS relaxation,} also for \PGO. 
 Currently, there are no certifiably optimal non-minimal solvers for shape alignment; 
 Zhou~\etal~\cite{Zhou17pami-shapeEstimationConvex} propose a convex relaxation to obtain an approximate solution for \shape.

\subsection{Robust Estimation}
\label{sec:rw_robust}

{\bf Global Methods.} 
{We refer to a robust method as \textit{global}, if it does not require an initial guess. 
Several global methods adopt the framework of \emph{consensus maximization}~\cite{Chin18ECCV-robustFittingMaxCon,Chin17slcv-maximumConsensusAdvances}, which looks for an estimate  that maximizes the number of measurements that are explained within a prescribed estimation error. Consensus maximization is NP-hard~\cite{Chin18ECCV-robustFittingMaxCon,Tzoumas19iros-outliers}, and 
related work investigates both approximations and exact algorithms.
\ransac~\cite{Fischler81} is a widely used \isExtended{heuristic~\cite{Meer91ijcv-robustVision,Hartley04book},}{heuristic,} applicable when a \textit{minimal} solver exists.
But \ransac provides no optimality guarantees, and its running time grows exponentially with the outlier ratio~\cite{Bustos18pami-GORE}. 
 Tzoumas~\etal~\cite{Tzoumas19iros-outliers} develop the general-purpose \emph{Adaptive Trimming} (\adapt) algorithm, that has linear running time, and, instead, is applicable when a \textit{non-minimal} solver exists.
Mangelson~\etal~\cite{Mangelson18icra} propose a graph-theoretic method to prune outliers in \PGO.
Exact solutions for consensus maximization are based on
BnB~\cite{Chin17slcv-maximumConsensusAdvances}:
 see~\cite{Bazin12accv-globalRotSearch}~for the Wahba problem, and~\cite{Bustos18pami-GORE} 
 for \PCR. 
} 

Another framework for global methods is \emph{M-estimation}, which resorts to robust cost functions.
Enqvist~\etal~\cite{Enqvist12eccv-robustFitting} use a \textit{truncated least squares} (TLS) cost, and propose an approach that scales, however, exponentially with
the dimension of the parameter space. 
More recently, \emph{SDP relaxations} have also been used to optimize robust costs.  
 Carlone and Calafiore~\cite{Carlone18ral-robustPGO2D} develop convex relaxations for \PGO with $\ell_1$-norm and Huber loss functions. Lajoie~\etal~\cite{Lajoie19ral-DCGM} adopt a TLS cost for \PGO. 
Yang and Carlone~\cite{Yang19rss-teaser,Yang19iccv-QUASAR} develop an SDP relaxation for the Wahba and \PCR problem, also adopting a TLS cost. Currently, the poor scalability of the state-of-the-art SDP solvers limits these algorithms to only small-size problems.  

Finally, \textit{graduated non-convexity} (\GNC) methods have also been employed to optimize robust costs~\cite{Blake1987book-visualReconstruction}. 
Rather than directly optimizing a non-convex robust cost, 
these methods 
sequentially optimize a sequence of surrogate functions, which start from a convex approximation of the original cost, but then gradually become non-convex, converging eventually to the original cost. 
Despite \GNC's success in early computer vision applications~\cite{Black96ijcv-unification,Gold98}, its broad applicability has remain limited due to the lack of {non-minimal} solvers. Indeed, only a few specialized methods for spatial perception have used \GNC.
Particularly, Zhou~\etal~\cite{Zhou16eccv-fastGlobalRegistration} develop a method for \PCR, {using Horn's or Arun's methods~\cite{Horn87josa,Arun87pami}}.

{\bf Local Methods.} In contrast to global methods, local methods require an initial guess. 
In the context of M-estimation, these methods iteratively optimize a robust cost function till they converge to a local minimum~\cite{Bosse17fnt}.
 Zach \etal~\cite{Zach14eccv} include auxiliary variables and propose an 
 iterative  optimization approach that alternates updates on the estimates and the auxiliary variables; 
 the approach still requires an initial guess.
Bouaziz~\etal~\cite{Bouaziz13acmsig-sparseICP} propose  
robust variants of the \emph{iterative closest point} algorithm for \PCR. 
S\"{u}nderhauf and Protzel~\cite{Sunderhauf12iros}, Olson and Agarwal~\cite{Olson12rss}, Agarwal~\etal~\cite{Agarwal13icra}, Pfingsthorn and Birk~\cite{Pfingsthorn16ijrr}
 propose local methods for \PGO. 
 Wang~\etal~\cite{Wang14cvpr-robustShapeEstimationL1} 
 investigate local methods for shape alignment.

\section{Black-Rangarajan Duality and Graduated Non-convexity}

We review the Black-Rangarajan duality~\cite{Black96ijcv-unification}, and a tool for global optimization known as graduated non-convexity~\cite{Blake1987book-visualReconstruction}. 


\subsection{Black-Rangarajan Duality}
\label{sec:BRduality}

This section revisits the Black-Rangarajan duality between robust estimation and outlier process~\cite{Black96ijcv-unification}. This theory is less known in robotics, and its applications have been mostly
targeting early vision problems, with few notable exceptions. 


\begin{lemma}[Black-Rangarajan Duality~\cite{Black96ijcv-unification}]
\label{lem:BRduality}
Given a robust cost function $\rho(\cdot)$, define $\phi(z) \doteq \rho(\sqrt{z})$. 
If $\phi(z)$ satisfies $\lim_{z\rightarrow 0} \phi'(z) = 1$, $\lim_{z\rightarrow \infty} \phi'(z) = 0$, and $\phi''(z) < 0$, then the robust estimation problem~\eqref{eq:generalEstimation}
is equivalent to 
\bea \label{eq:optiOutlierProcess} 
\min_{\vxx \in \calX, w_i \in [0,1]} \sumAllPointsi  \left[ w_i r^2(\vy_i,\vxx) + \Phi_\rho(w_i) \right],
\eea
where $ w_i \in [0,1]$ ($i=1,\dots,N$) are slack variables (or \emph{weights}) associated to each measurement $\vy_i$, 
and the 
  function $\Phi_\rho(w_i)$ (the so called \emph{outlier process}) defines a penalty on the weight $w_i$. 
  The  expression of $\Phi_\rho(w_i)$ depends on the choice of robust cost function $\rho(\cdot)$. 
\end{lemma}

The conditions on $\rho(\cdot)$ are satisfied by all common choices of robust costs~\cite{Black96ijcv-unification}.  
Besides presenting this fundamental result, asserting the equivalence between the outlier process~\eqref{eq:optiOutlierProcess}  and the robust estimation~\eqref{eq:generalEstimation}, 
Black and Rangarajan provide a procedure (see Fig.~10 in~\cite{Black96ijcv-unification}) to compute $\Phi_\rho(w_i)$ 
and show that common robust cost functions admit a simple analytical expression for $\Phi_\rho(w_i)$. 
Interestingly, the outlier process~\eqref{eq:optiOutlierProcess} has been often used in robotics and SLAM~\cite{Agarwal13icra,Sunderhauf12iros}, without acknowledging the connection with robust estimation, and 
 with a heuristic design of the penalty terms $\Phi_\rho(w_i)$. 

Despite the elegance of~\prettyref{lem:BRduality}, Problem~\eqref{eq:optiOutlierProcess} remains hard to solve, 
due to its non-convexity. 
Approaches in robotics (\eg~\cite{Agarwal13icra,Sunderhauf12iros}) apply local optimization from an initial guess, 
resulting in brittle solutions (see~\cite{Carlone18ral-robustPGO2D} and Section~\ref{sec:applications}). 
\isExtended{To date, the only provably optimal solvers for~\eqref{eq:optiOutlierProcess} (when $\rho(\cdot)$ is a truncated least square cost) are given in~\cite{Yang19rss-teaser,Yang19iccv-QUASAR,Lajoie19ral-DCGM}.}{}



\subsection{Graduated Non-Convexity (\GNC)}
\label{sec:graduatedNonConvexity}

Graduated non-convexity (\GNC) is a popular approach for the optimization of a generic non-convex cost function $\rho(\cdot)$ and has been used in several endeavors, including vision~\cite{Blake1987book-visualReconstruction} and machine learning~\cite{Rose1998IEEE-deterministicAnnealing} (see~\cite{Mobahi2015WCVPR-continuationConvexEnvelope} for more applications).
 The basic idea of \GNC is to introduce 
 a surrogate cost $\rho_{\mu}(\cdot)$, governed by a control parameter $\mu$, such that (i) for a certain value of $\mu$, the function $\rho_{\mu}(\cdot)$ is convex, 
 and (ii) in the limit (typically for $\mu$ going to 1 or infinity) one recovers the original (non-convex) $\rho(\cdot)$. 
 Then \GNC computes a solution to the non-convex problem by starting from its convex surrogate and gradually changing $\mu$ (\ie gradually increasing the amount of non-convexity) till the original non-convex function is recovered. The solution obtained at each iteration is used as the initial guess for the subsequent iteration. 

 Let us shed some light on \GNC with two examples.
 \setcounter{theorem}{0}
  \begin{example}[Geman McClure (GM) and \GNC] 
  \label{example:GM}
  The Geman-McClure function is a popular (non-convex) robust cost. 
  The following equation shows the GM function (left) and the surrogate function  including 
  a control parameter $\mu$ (right):
  \bea \label{eq:formulationGNCGM}
\rho(r) \doteq \frac{\barcsq r^2}{ \barcsq +r^2}
  \;\; \Longrightarrow \;\;
 \rho_{\mu}(r) = \frac{\mu \barcsq r^2}{ \mu \barcsq + r^2},
\eea
where $\barc$ is a given parameter that 
 determines the shape of the Geman McClure function $\rho(r)$. 

The surrogate function $\rho_{\mu}(r)$ (shown in Fig.~\ref{fig:GNCplots}(a)) is such that: (i)  $\rho_{\mu}(r)$ becomes convex for large $\mu$ (in the limit of $\mu \rightarrow \infty$, $\rho_{\mu}(r)$ becomes quadratic), and (ii) $\rho_{\mu}(r)$ recovers $\rho(r)$  when $\mu = 1$. 
\GNC minimizes the function $\rho(r)$ by repeatedly minimizing the function $\rho_{\mu}(r)$ for decreasing values of $\mu$. 
\end{example}

\begin{example}[Truncated Least Squares (TLS) and \GNC] 
\label{example:TLS}
The truncated least squares function is defined as:
\bea
\label{eq:formulationTLS}
\hspace{-2mm} \rho(r) = 
\begin{cases}
\scriptstyle{ r^2 } & {\footnotesize\text{ if }}  \scriptstyle{ r^2 \in \left[0, \barcsq \right] } \\
 \scriptstyle{ \barcsq } & {\footnotesize\text{ if }}  \scriptstyle{ r^2 \in \left[\barcsq, +\infty \right) }
\end{cases},
\eea
where $\barc$ is a given truncation threshold. 
The \GNC surrogate function with control parameter $\mu$ is:
\bea
\hspace{-0mm} \rho_\mu(r) = \begin{cases}
\scriptstyle{ r^2 } & {\footnotesize\text{ if }} \scriptstyle{ r^2 \in \left[0, \frac{\mu }{\mu + 1}\barcsq \right] } \\
  \scriptstyle{ 2\barc | r | \sqrt{\mu(\mu+1)} - \mu (\barcsq + r^2)  } & {\footnotesize\text{ if }} \scriptstyle{ r^2 \in \left[\frac{\mu }{\mu + 1} \barcsq, \frac{\mu+1}{\mu}\barcsq \right] }\\
 \scriptstyle{ \barcsq } & {\footnotesize\text{ if }} \scriptstyle{ r^2 \in \left[\frac{\mu+1}{\mu}\barcsq, +\infty \right) }
\end{cases}.
\eea
By inspection, one can verify $\rho_\mu(r)$ is convex for $\mu$ approaching zero ($\rho''_{\mu}(r) = - 2\mu \rightarrow 0$) and retrieves $\rho(r)$  in~\eqref{eq:formulationTLS} for $\mu\rightarrow+\infty$.
An illustration of $\rho_\mu(r)$ 
is given in Fig.~\ref{fig:GNCplots}(b).
\end{example}


\newcommand{\mpw}{4.7cm}
\begin{figure}[t!]
	\begin{center}
	\begin{minipage}{\columnwidth}
	\hspace{-0.2cm}
	\begin{tabular}{cc}%
		\hspace{-5mm}
			\begin{minipage}{\mpw}%
			\centering%
			\includegraphics[width=1.0\columnwidth]{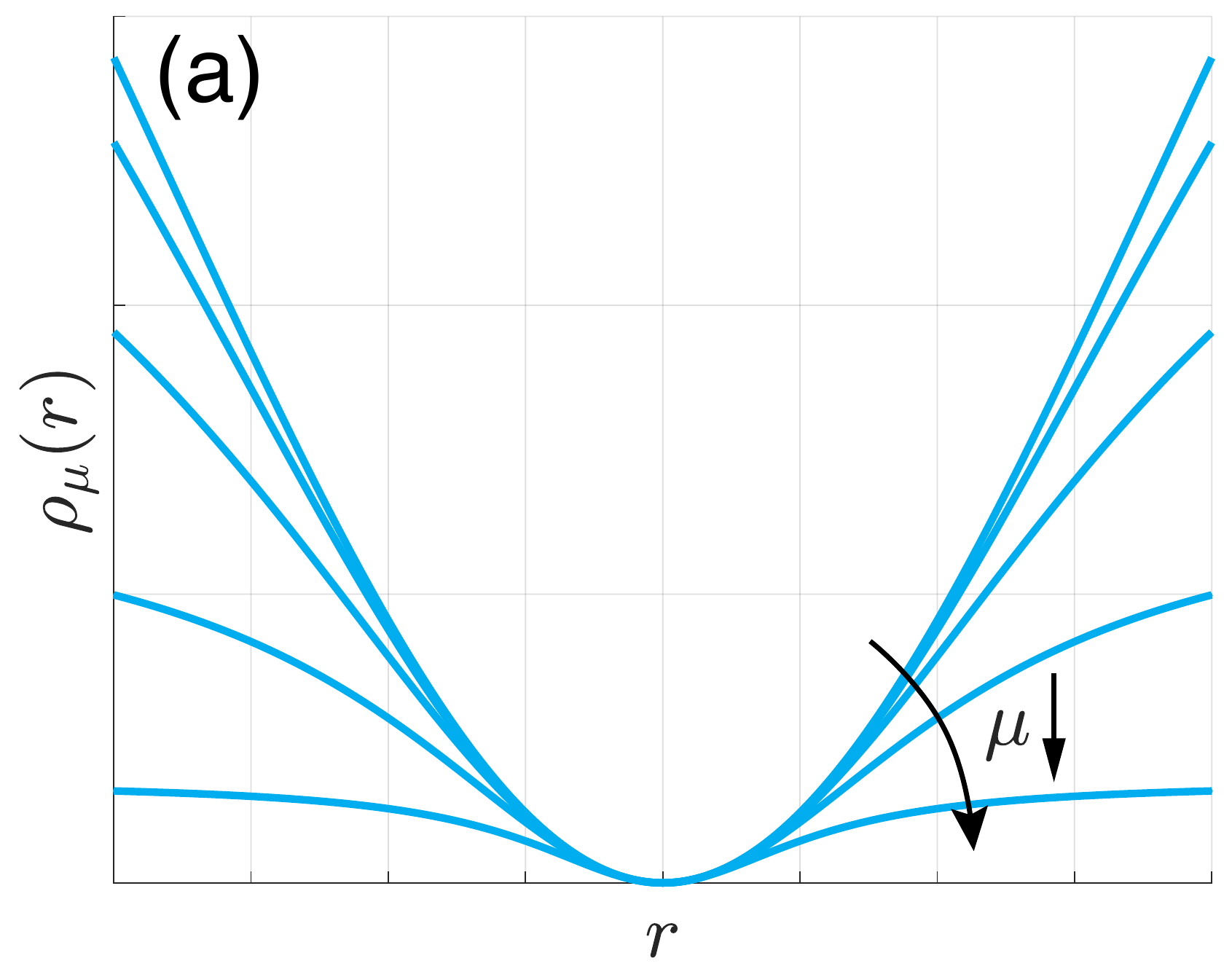}
			\end{minipage}
		& \hspace{-7mm}
			\begin{minipage}{\mpw}%
			\centering%
			\includegraphics[width=1.0\columnwidth]{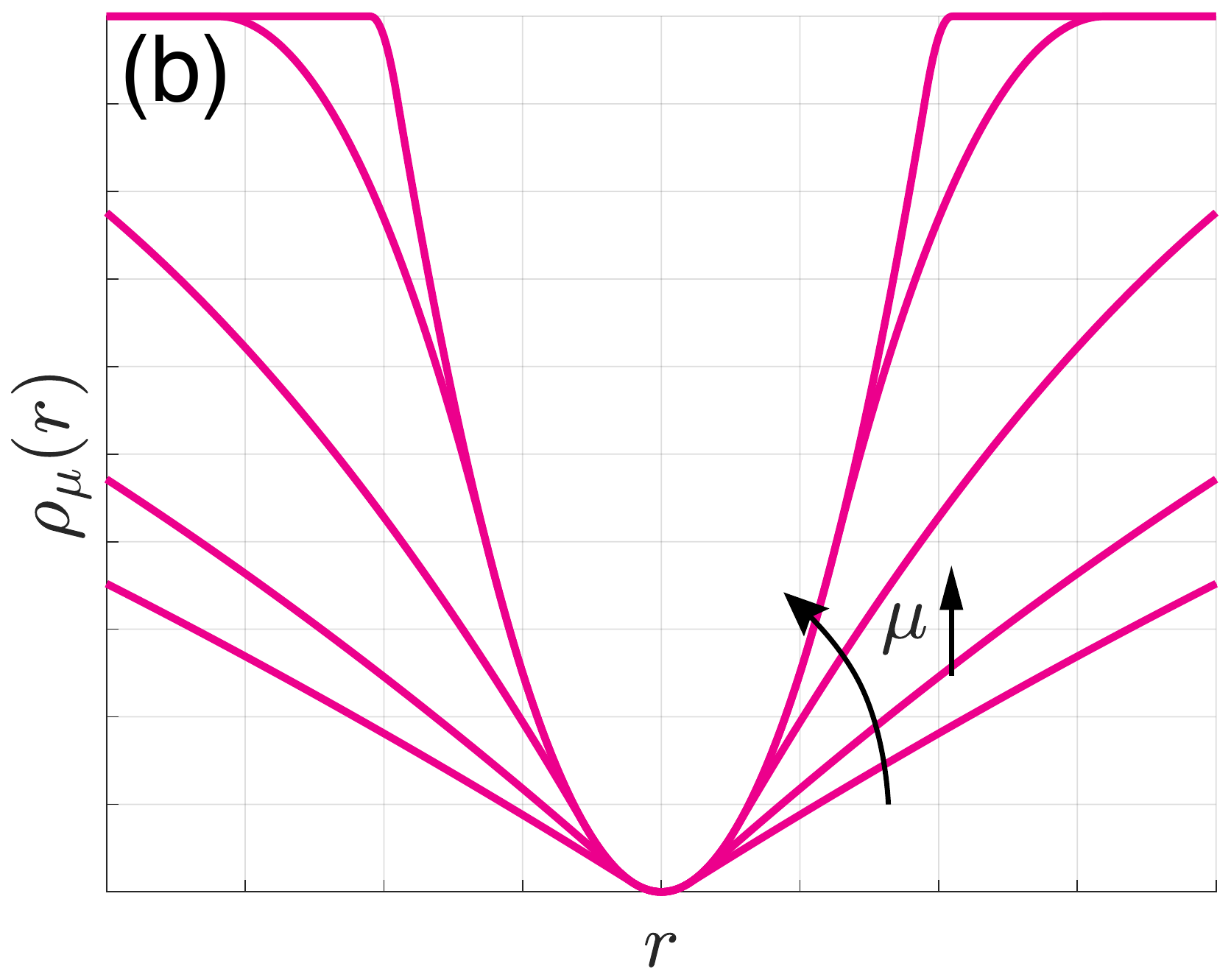}
			\end{minipage}
		\end{tabular}
	\end{minipage}
	\begin{minipage}{\textwidth}
	\end{minipage}
	\vspace{-4mm}
	\caption{Graduated Non-Convexity (\GNC) with control parameter $\mu$ for 
	 (a) Geman McClure (GM) and (b) Truncated Least Squares (TLS) costs.
	\label{fig:GNCplots}}
	\vspace{-8mm} 
	\end{center}
\end{figure}


\section{The \GNC Algorithm for Robust Estimation}
\label{sec:GNCalgorithm}

We present an algorithm that combines \GNC, Black-Rangarajan duality, and non-minimal solvers to 
solve\revise{~the robust estimation} problem~\eqref{eq:generalEstimation} without an initial guess. 


\subsection{Overview: \GNC Algorithm with Non-minimal Solvers} 
\label{sec:GNC-algorithm-overview}

We start by providing an overview of the proposed algorithm, then we delve into technical details and 
tailor the approach to two specific robust cost functions (Section~\ref{sec:GNC-GM-TLS}).  
Instead of optimizing directly\revise{~the robust estimation problem}~\eqref{eq:generalEstimation}, we use \GNC and, 
at each \emph{outer} iteration, we fix a $\mu$ and optimize: 
\bea \label{eq:GNCestimation}
\min_{\vxx \in \calX} \sumAllPointsi \rho_\mu( r(\vy_i,\vxx) ).
\eea 
Since non-minimal solvers cannot be used directly to solve~\eqref{eq:GNCestimation}, 
we use the Black-Rangarajan duality and rewrite~\eqref{eq:GNCestimation} using the corresponding outlier process:
\bea \label{eq:GNC-outlierProcess}
\min_{\vxx \in \calX, w_i \in [0,1]} \sumAllPointsi  \left[ w_i r^2(\vy_i,\vxx) + \Phi_{\rho_\mu}(w_i) \right].
\eea 
As discussed in Section~\ref{sec:GNC-GM-TLS} and~\cite{Black96ijcv-unification}, it is easy to compute the penalty terms $\Phi_{\rho_\mu}(\cdot)$ even for the surrogate function $\rho_\mu(\cdot)$. 

Finally, we solve~\eqref{eq:GNC-outlierProcess} by alternating optimization, where  
 at each \emph{inner} iteration we first optimize over the $\vxx$ (with fixed $w_i$), and 
 then we optimize over $w_i$, $i=1,\ldots,N$ (with fixed $\vxx$).  
 In particular, at inner iteration $t$, we perform the following:

\begin{enumerate}
\item {\bf Variable update}: minimize~\eqref{eq:GNC-outlierProcess} with respect to $\vxx$ with fixed weights $w\at{t-1}_i$:
\bea \label{eq:variableUpdate}
\vxx\at{t} = \argmin_{\vxx \in \calX} \sumAllPointsi w\at{t-1}_i r^2(\vy_i, \vxx),
\eea
where we dropped the second term in~\eqref{eq:GNC-outlierProcess} which does not depend on $\vxx$. 
Problem~\eqref{eq:variableUpdate} is simply a weighted version of the outlier-free problem~\eqref{eq:leastSquares}, 
 hence it can be solved globally using certifiably optimal non-minimal solvers.  

\item {\bf Weight update}: minimize~\eqref{eq:GNC-outlierProcess} with respect to $w_i$ ($i=1,\ldots,N$) 
with fixed $\vxx\at{t}$:
\bea \label{eq:weightUpdate}
\vw\at{t} = \argmin_{w_i \in [0,1]} \sumAllPointsi \left[ w_i r^2(\vy_i, \vxx\at{t})  + \Phi_{\rho_\mu}(w_i)\right],
\eea
where $r^2(\vy_i, \vxx\at{t})$ is a constant for fixed $\vxx$, and the expression of $\Phi_{\rho_\mu}(\cdot)$ depends on the choice of robust cost function. As we discuss in Section~\ref{sec:GNC-GM-TLS}, the weight update can be typically solved in closed form. 
\end{enumerate}

The process is then repeated for changing values of $\mu$, where each change of $\mu$ increases the amount of non-convexity. 
\setcounter{theorem}{1}
\begin{remark}[Teaching an Old Dog New Tricks]
While the combination of \GNC and Black-Rangarajan duality has been investigated in related works~\cite{Black96ijcv-unification,Zhou16eccv-fastGlobalRegistration}, its applicability has been limited by the lack of global solvers for the variable update in~\eqref{eq:variableUpdate}. 
For instance,~\cite{Zhou16eccv-fastGlobalRegistration} focuses on a specific problem (point cloud registration) where~\eqref{eq:variableUpdate} can be solved in closed form~\cite{Arun87pami,Horn87josa},  
while~\cite{Black96ijcv-unification} focuses on a Markov Random Field formulation for which global solvers and heuristics exist~\cite{Hu19ral-fuses}.
 One of the main insights behind our approach is that modern non-minimal solvers (developed over the last 5 years) 
 allow solving~\eqref{eq:variableUpdate} globally for a broader class of problems,  
 including spatial perception problems such as SLAM, mesh registration, and object localization from images.
\end{remark}

\isExtended{
\begin{remark}[Generality]
The algorithm described in this section
is applicable to any estimation problem and robust 
cost function as long as: (i) we can develop a surrogate function $\rho_\mu$ for \GNC and find an analytic penalty function $\Phi_{\rho_\mu}$, 
(ii) we have an efficient way to solve the variable update~\eqref{eq:variableUpdate}, and 
(iii) we have an efficient way to solve the weight update~\eqref{eq:weightUpdate}.
Condition (i) is relatively mild and~\cite{Black96ijcv-unification} already provides analytic expressions for the most common robust costs. Condition (iii) is also mild since the optimization splits into $N$ scalar optimizations, whose solution can be typically computed in closed form. 
Therefore, the only ``stringent'' requirement is the availability of non-minimal solver for~\eqref{eq:variableUpdate}. 
Luckily, non-minimal solvers exist for several key spatial perception applications as the ones considered in Section~\ref{sec:applications}. 
\end{remark}
}{}

\newcommand{\mpwthree}{6cm}
\newcommand{\myhspace}{\hspace{-3mm}}

\begin{figure*}[h!]
	\begin{center}
	\begin{minipage}{\textwidth}
	\begin{tabular}{ccc}%
		\hspace{-4mm}
			\begin{minipage}{\mpwthree}%
			\centering%
			\includegraphics[width=\columnwidth]{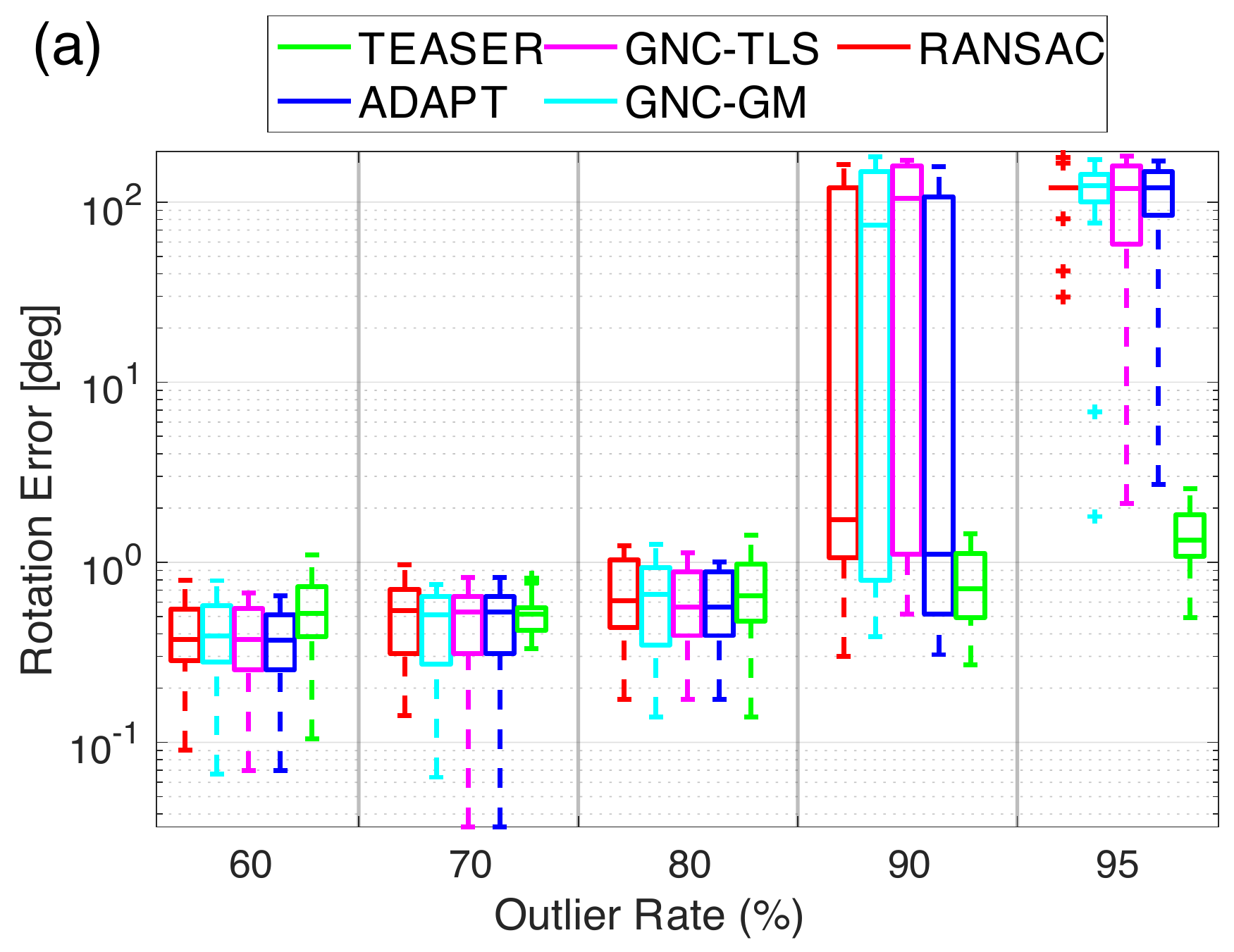} \\
			\end{minipage}
		& \myhspace
			\begin{minipage}{\mpwthree}%
			\centering%
			\includegraphics[width=\columnwidth]{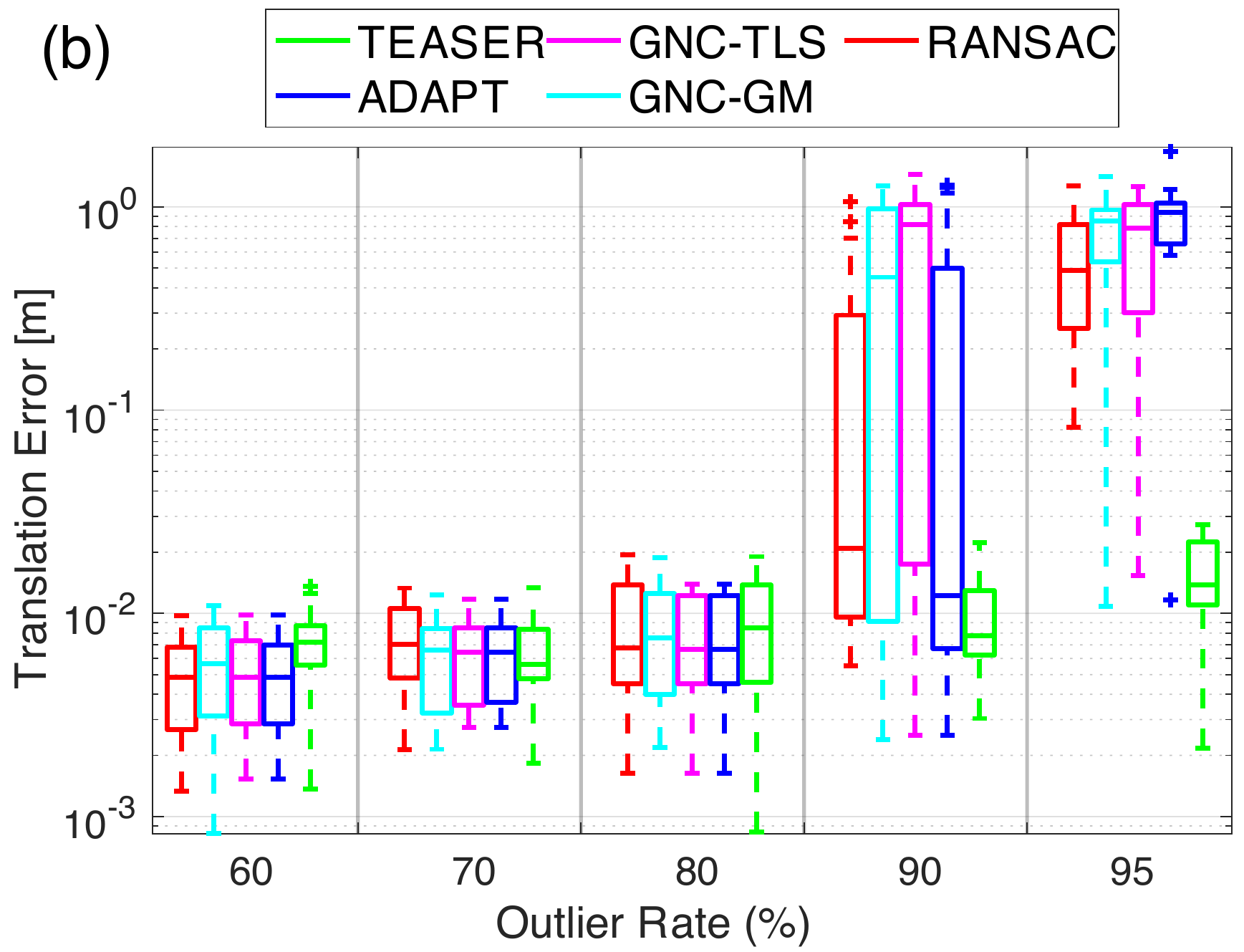} \\
			\end{minipage}
		& \hspace{-5mm}
			\begin{minipage}{\mpwthree}%
			\centering%
			\includegraphics[width=\columnwidth]{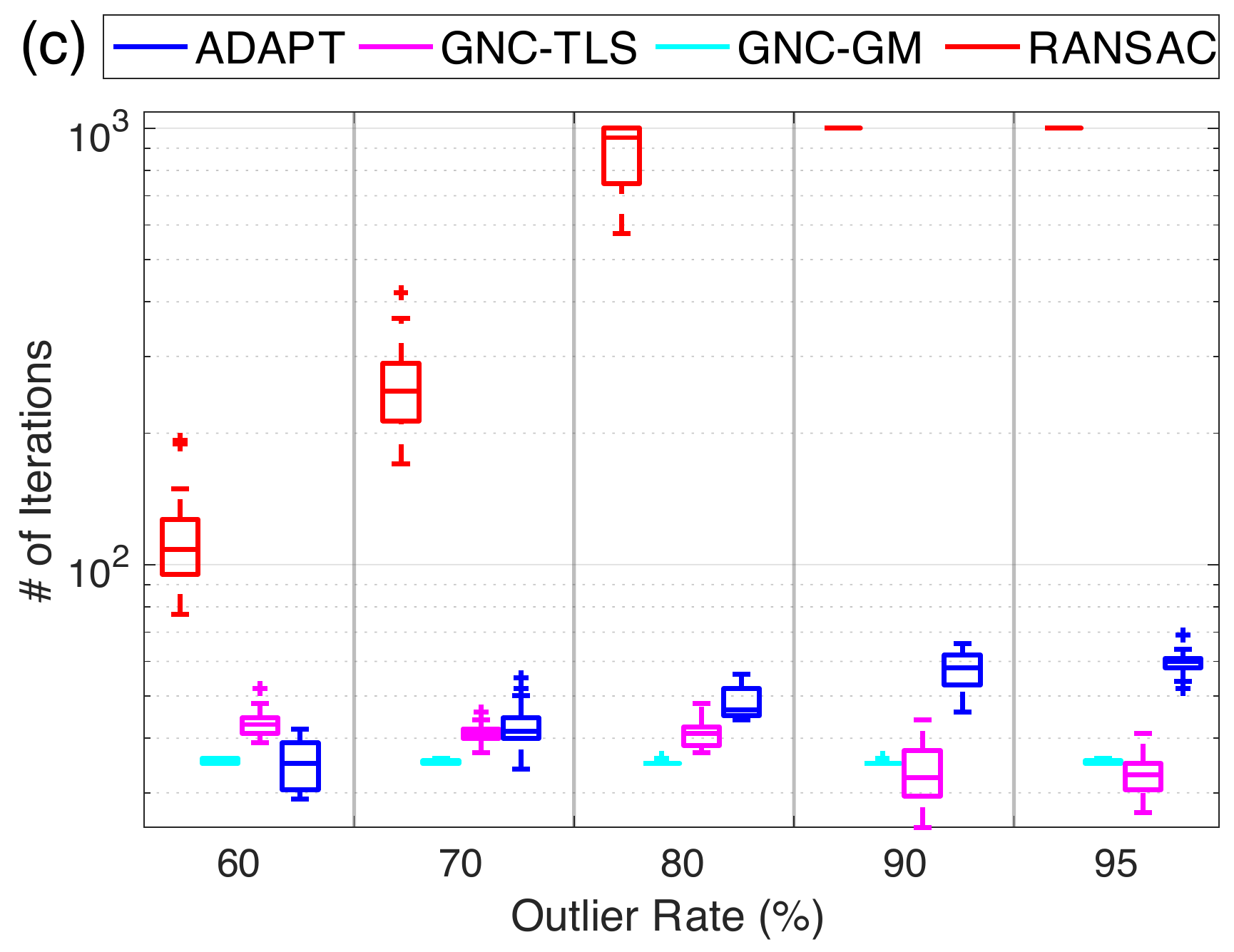} \\
			\end{minipage}
		\end{tabular}
	\end{minipage}
	\vspace{-3mm} 
	\caption{ {\bf Point Cloud Registration.} Performance of \GNCGM and \GNCTLS compared with state-of-the-art techniques on the \bunny dataset~\cite{Curless96siggraph} for increasing outliers. 
	(a) rotation error; (b) translation  error; (c) number of iterations until convergence. Statistics are computed over 20 Monte Carlo runs.
	}
	\label{fig:pointCloudRegistration}
	\vspace{-5mm} 
	\end{center}
\end{figure*}


\begin{figure*}[h!]
	\begin{center}
	\begin{minipage}{\textwidth}
	\begin{tabular}{ccc}%
		\hspace{-4mm}
			\begin{minipage}{\mpwthree}%
			\centering%
			\includegraphics[width=\columnwidth]{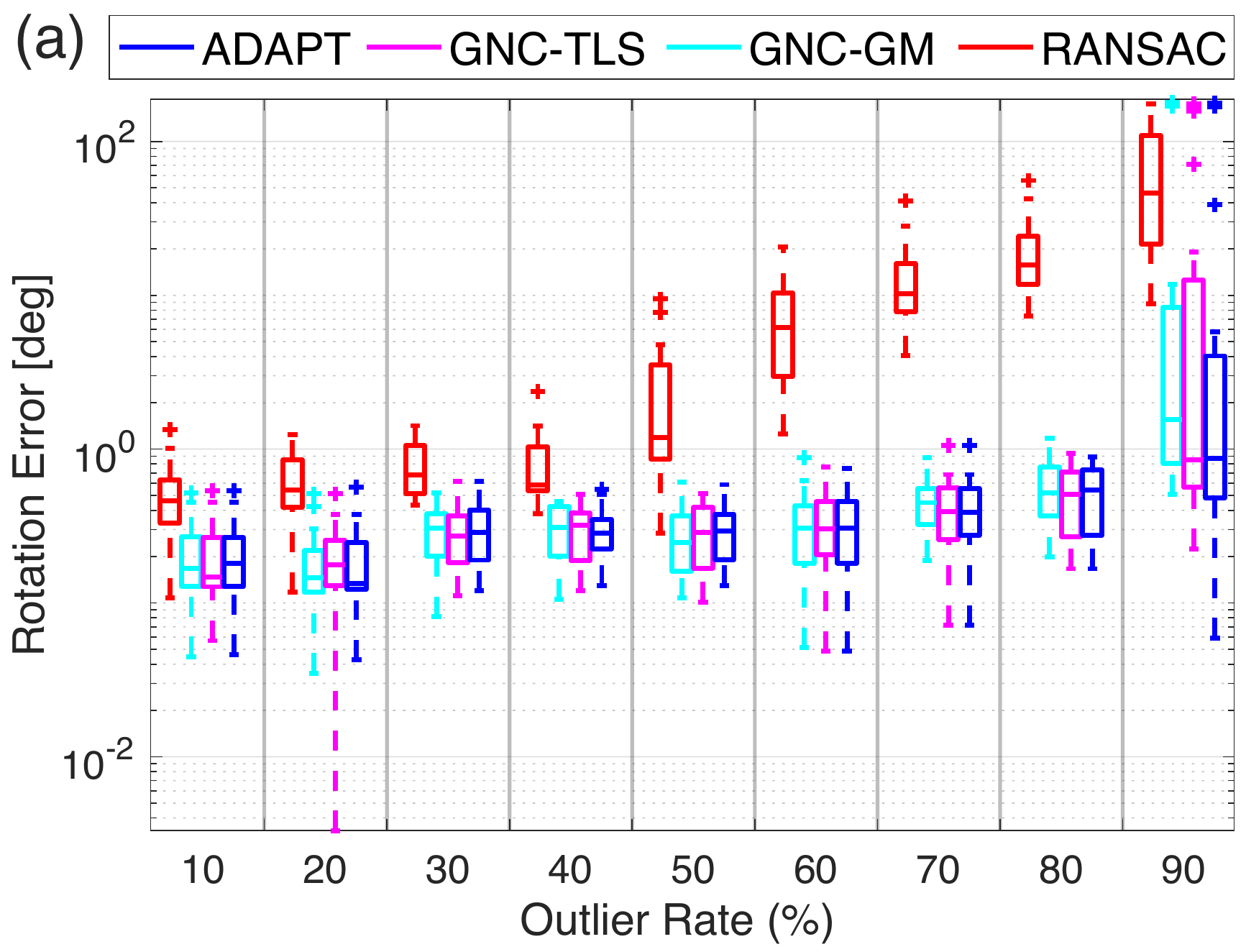} \\
			\end{minipage}
		& \myhspace
			\begin{minipage}{\mpwthree}%
			\centering%
			\includegraphics[width=\columnwidth]{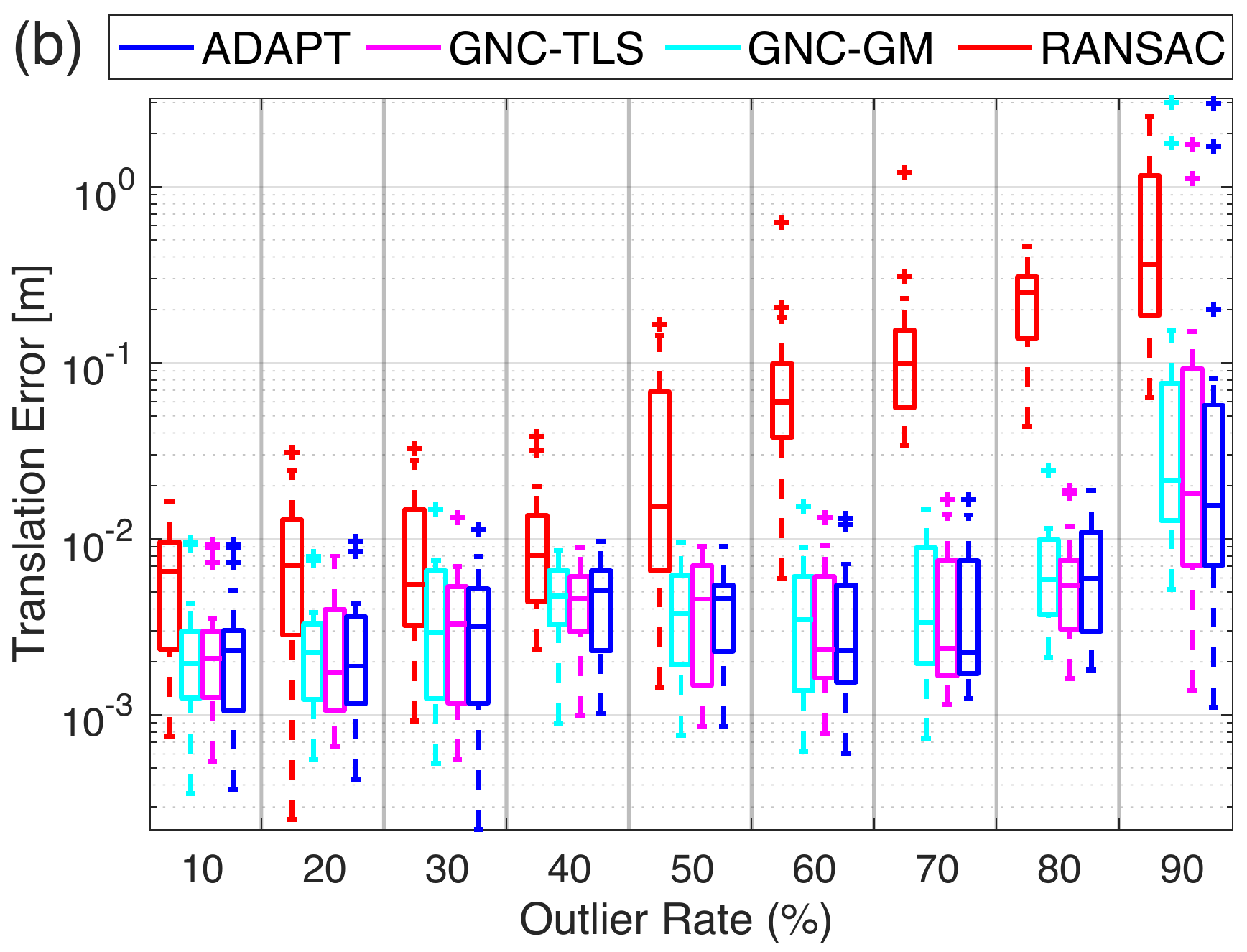} \\
			\end{minipage}
		& \hspace{-5mm}
			\begin{minipage}{\mpwthree}%
			\centering%
			\includegraphics[width=\columnwidth]{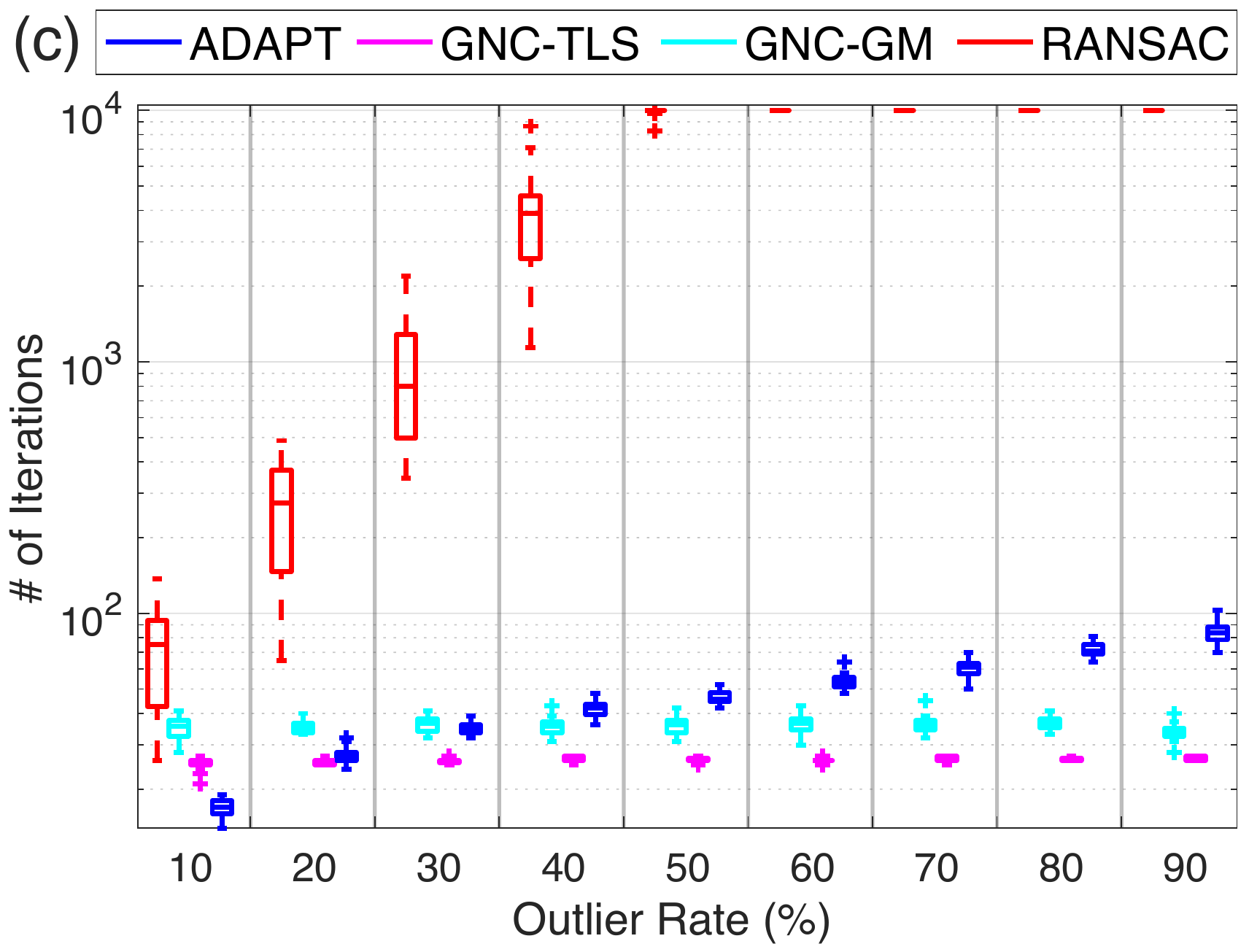} \\
			\end{minipage} \\ 
		\hspace{-4mm}
			\begin{minipage}{\mpwthree}%
			\centering%
			\includegraphics[width=\columnwidth]{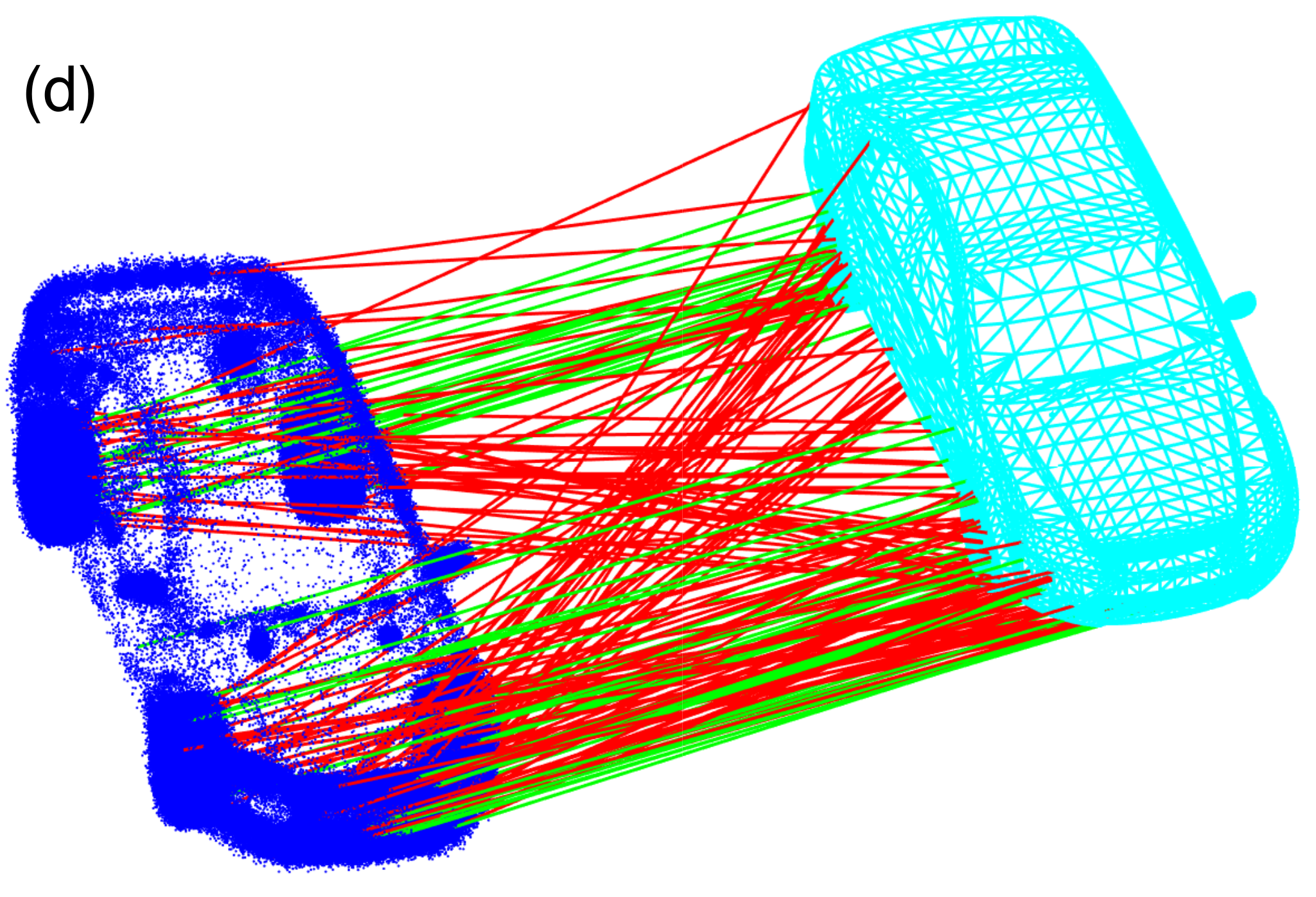} \\
			\end{minipage}
		& \myhspace
			\begin{minipage}{\mpwthree}%
			\centering%
			\includegraphics[width=\columnwidth]{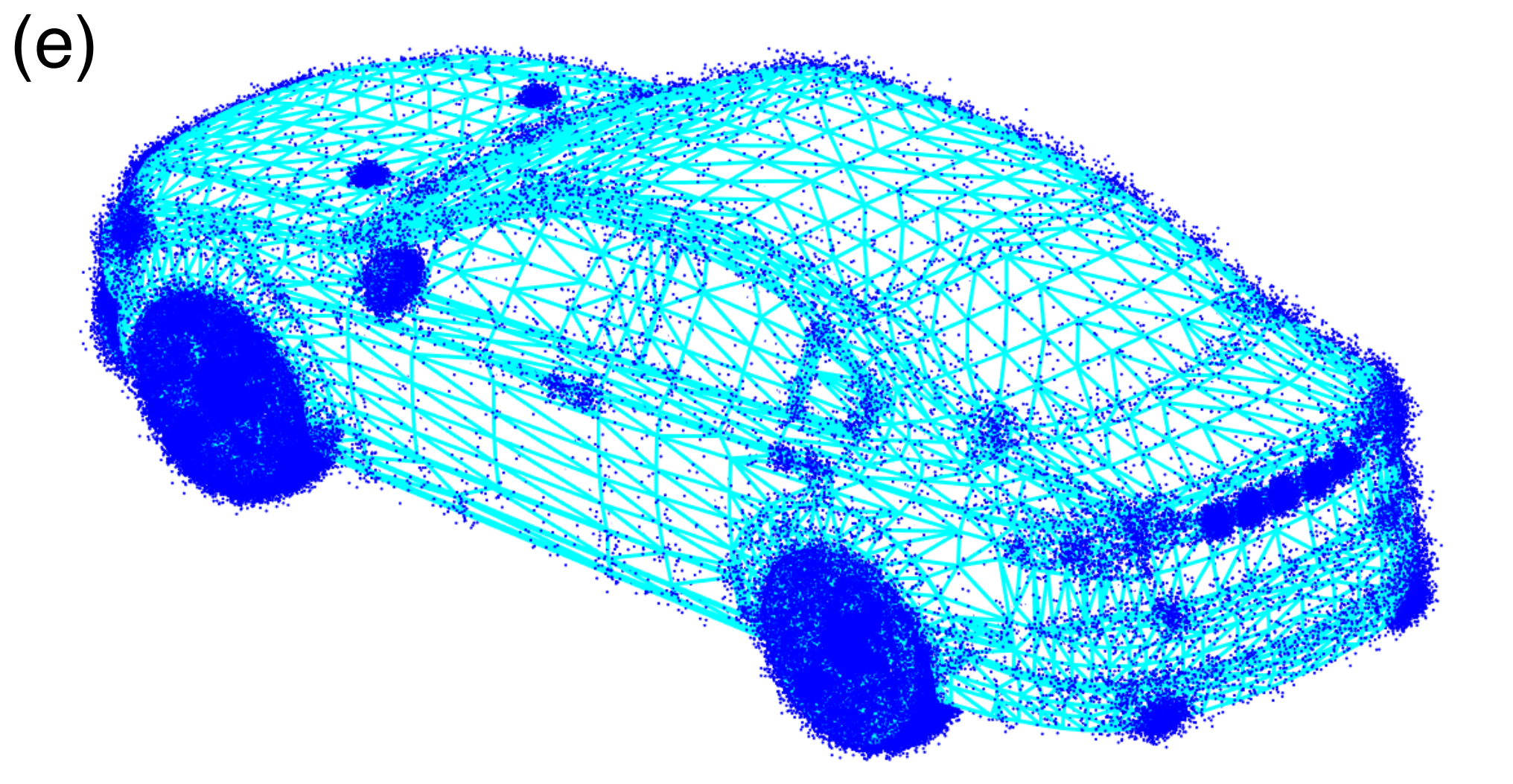} \\
			\end{minipage}
		& \myhspace
			\begin{minipage}{\mpwthree}%
			\centering%
			\includegraphics[width=\columnwidth]{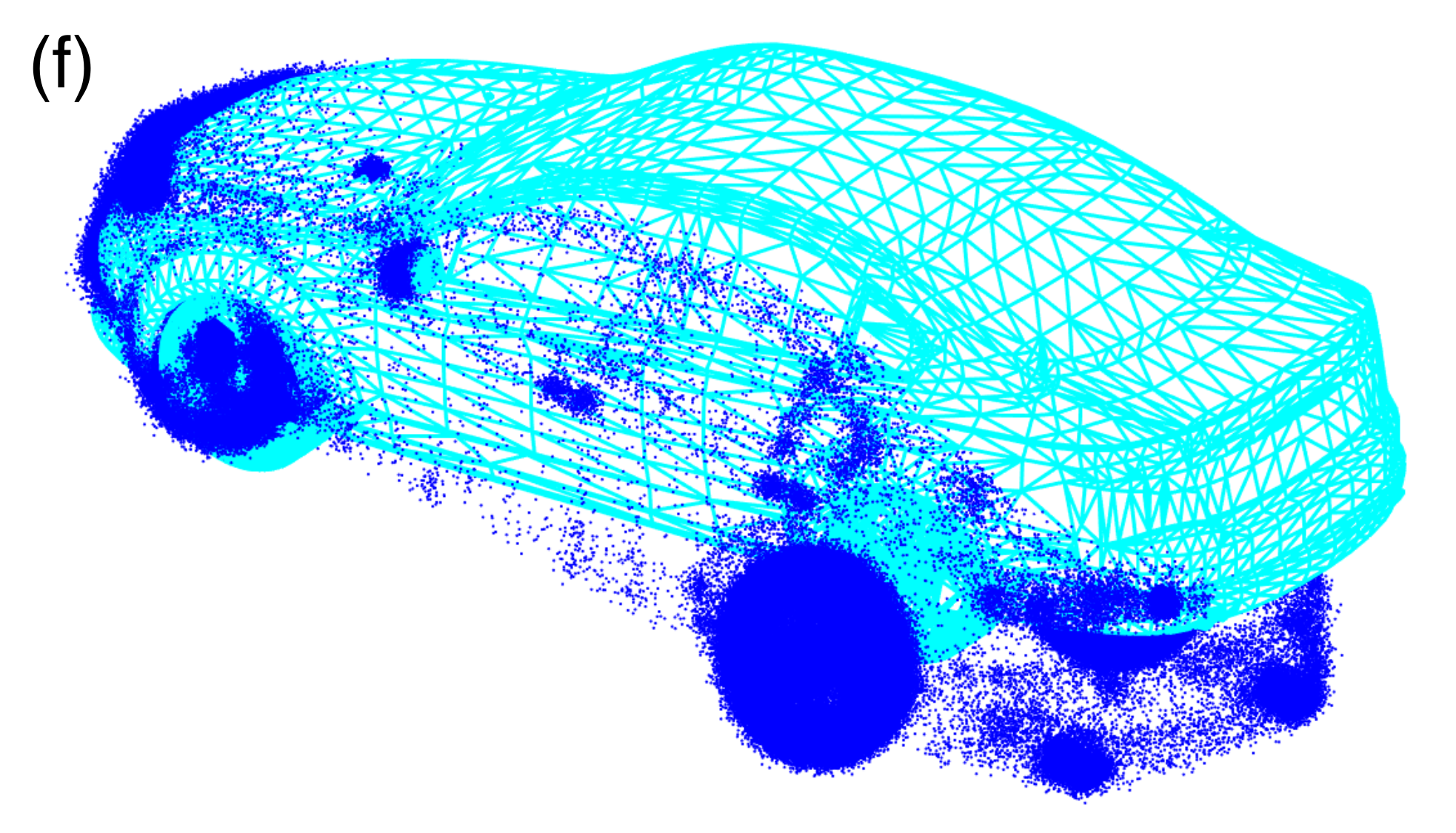} \\
			\end{minipage} \\
	\end{tabular}
	\end{minipage}
	\vspace{-3mm} 
	\caption{{\bf Mesh Registration.} Performance of \GNCGM and \GNCTLS compared with state-of-the-art techniques on mesh registration on the \PASCALplus ``car-2'' dataset~\cite{Xiang2014WACV-PASCAL+} for increasing outliers.
     {
     Quantitative results: (a) rotation error; (b) translation error; (c) number of iterations until convergence.
     Qualitative results: (d)  point cloud and mesh with putative correspondences (70\% outliers; green: inliers, red: outliers); (e) successful registration using \GNCTLS; 
     (f) incorrect registration using \ransac.
     }
	Statistics are computed over 20 Monte Carlo runs.
	\label{fig:meshRegistration}}
	\vspace{-7.5mm} 
	\end{center}
\end{figure*}


\subsection{The \GNC Algorithm with GM and TLS Costs}
\label{sec:GNC-GM-TLS}

Here we tailor the \GNC algorithm to two cost functions, the 
Geman McClure and the Truncated Least Squares costs, provide expressions for 
the penalty term $\Phi_{\rho_\mu}$, and discuss how to solve the weight update step. 
The proofs of the following propositions are given in the supplementary material~\cite{Yang20tr-GNC}. 

\begin{proposition}[GNC-Geman McClure (\GNCGM)] \label{prop:GNCGM}
Consider the Geman-McClure function and its \GNC surrogate with control parameter $\mu$, as introduced in Example~\ref{example:GM}. Then, the minimization of the surrogate function $\rho_\mu(\cdot)$ is equivalent to  the 
outlier process with penalty term chosen as:
\bea 
\label{eq:outlierProcessGNCGM}
\Phi_{\rho_\mu}(w_i) = \mu \barcsq (\sqrt{w_i} - 1)^2.
\eea
Moreover, defining the residual $\hatr_i^2 \doteq r^2(\vy_i, \vxx\at{t})$, 
 the weight update at iteration $t$ can be solved in closed form as:
\bea 
\label{eq:dualWeightUpdateGNCGM}
w_i\at{t} = \left( \frac{\mu \barcsq }{\hatr_i^2 + \mu \barcsq} \right)^2\!.
\eea
\end{proposition}


\begin{proposition}[GNC-Truncated Least Squares (\GNCTLS)] \label{prop:GNCTLS}
Consider the truncated least squares function and its \GNC surrogate with control parameter $\mu$, as introduced in Example~\ref{example:TLS}. Then, the minimization of the surrogate function $\rho_\mu(\cdot)$ is equivalent to  the 
outlier process with penalty term:
\bea \label{eq:outlierProcessGNCTLS}
\Phi_{\rho_\mu}(w_i) = \frac{\mu (1-w_i)}{ \mu + w_i} \barcsq.
\eea
Moreover, defining the residual $\hatr_i^2 \doteq r^2(\vy_i, \vxx\at{t})$, 
 the weight update at iteration $t$ can be solved in closed form as:
\bea \label{eq:dualWeightUpdateGNCTLS}
w_i\at{t} = \begin{cases}
0 & \text{ if } \hatr_i^2 \in \left[ \frac{\mu+1}{\mu}\barcsq, +\infty \right] \\
\frac{\barc}{\hatr_i}\sqrt{\mu(\mu+1)} - \mu & \text{ if }  \hatr_i^2 \in \left[ \frac{\mu}{\mu+1}\barcsq ,\frac{\mu+1}{\mu}\barcsq \right] \\
1 &\text{ if }  \hatr_i^2 \in \left[ 0,\frac{\mu}{\mu+1}\barcsq \right].
\end{cases}
\eea
\end{proposition}

\begin{remark}[Implementation Details] 
For \GNCGM, we start with a convex surrogate ($\mu\kern-0.5em\rightarrow\kern-0.5em\infty$) and decrease $\mu$ till we recover the original cost ($\mu\!\rightarrow \!1$). In practice, 
calling $r_{\max}^2 \doteq \max_i( r^2(\vy_i, \vxx^{(0)}))$ the maximum residual after the first variable update,  we initialize $\mu \!= \!2r^2_{\max} / \barcsq$,
 update $\mu \leftarrow \mu / 1.4$ at each outer iteration, 
 and stop when $\mu$ decreases below $1$.
For \GNCTLS, we start with a convex surrogate ($\mu\rightarrow 0$) and increase the $\mu$ till we recover the original cost ($\mu\kern-0.2em\rightarrow\kern-0.2em\infty$). In practice, 
we initialize  $\mu \!= \!\barcsq / ( 2 r^2_{\max} - \barcsq )$,
 update $\mu\!\leftarrow \!1.4 \mu$ at each outer iteration, 
 and stop when the sum of the weighted residuals $\sumAllPointsi w_i \hatr_i^2$ converges.  
 For each outer iteration we perform a single variable and weight update.\revise{~At the first iteration, all weights are set to 1 ($w_i^{(0)}=1, i=1,\dots,N$).} For both robust functions, we set the parameter $\barc$ to the maximum {error expected for the inliers, see Remarks 1-2 in~\cite{Yang19iccv-QUASAR}.}
\end{remark}


\section{Applications and Experiments}
\label{sec:applications}

\isExtended{This section shows that the proposed robust non-minimal solvers 
are robust to {70-80\%} outliers, 
outperform \ransac, 
are more accurate than specialized \emph{local} solvers, and 
faster than specialized \emph{global} solvers. 
 We showcase our approach in 
three spatial perception applications: point and mesh registration (Section~\ref{sec:REG}),
 pose graph optimization (Section~\ref{sec:PGO}), and 
 shape alignment (Section~\ref{sec:SA}).}
 {
We showcase our robust non-minimal solvers in 
three spatial perception applications: point cloud and mesh registration (Section~\ref{sec:REG}),
 pose graph optimization (Section~\ref{sec:PGO}), and 
 shape alignment (Section~\ref{sec:SA}).
 }

\subsection{3D Point Cloud and Mesh Registration}
\label{sec:REG}

\myParagraph{Setup} In generalized 3D registration, given a set of 3D points 
$\va_i \in \Real{3}$, $i=1,\ldots,N$, 
and a set of primitives $\MP_i$, $i=1,\ldots,N$ 
(being points, lines and/or planes) with putative correspondences 
$\va_i \leftrightarrow \MP_i$ (potentially including outliers), 
the goal is to find the best 
rotation $\MR \in \SOthree$ and translation $\vt \in \Real{3}$
that align the point cloud to the 3D primitives. The residual error is 
$r(\MR,\vt) = d(\MP_i,\MR \va_i + \vt)$, 
where $d(\cdot)$ denotes the distance between a primitive $\MP_i$ and a 
point $\va_i$ after the transformation $(\vt,\MR)$ is applied. 
The formulation can also accommodate weighted distances to account for heterogeneous and anisotropic measurement noise. 
In the outlier-free case, Horn's method~\cite{Horn87josa} gives a closed-form solution when all the 3D primitives are points and the noise is isotropic, and~\cite{Briales17cvpr-registration} develops a certifiably optimal relaxation when the 3D primitives include points, lines, and planes and the noise is anisotropic. 
We now show that, using the \GNCGM and \GNCTLS solvers,
we can efficiently robustify these non-minimal solvers. 
We benchmark our algorithms against state-of-the-art techniques in point cloud registration and mesh registration.



\begin{figure*}[t!] 
	\begin{center}
	\begin{minipage}{\textwidth}
	\begin{tabular}{ccc}%
		\hspace{-4mm}
			\begin{minipage}{\mpwthree}%
			\centering%
			\includegraphics[width=\columnwidth]{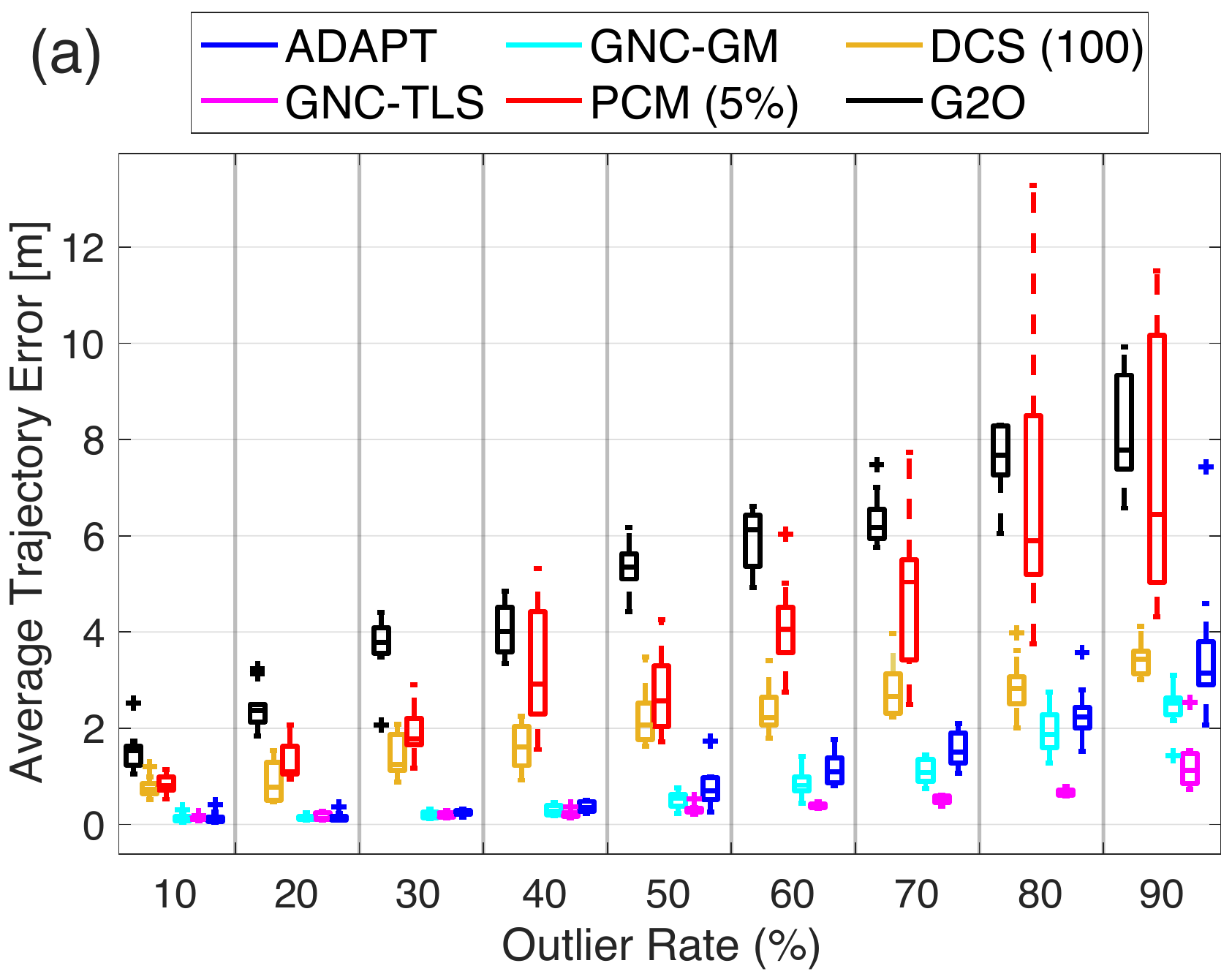} \\
			\end{minipage}
		& \myhspace
			\begin{minipage}{\mpwthree}%
			\centering%
			\includegraphics[width=\columnwidth]{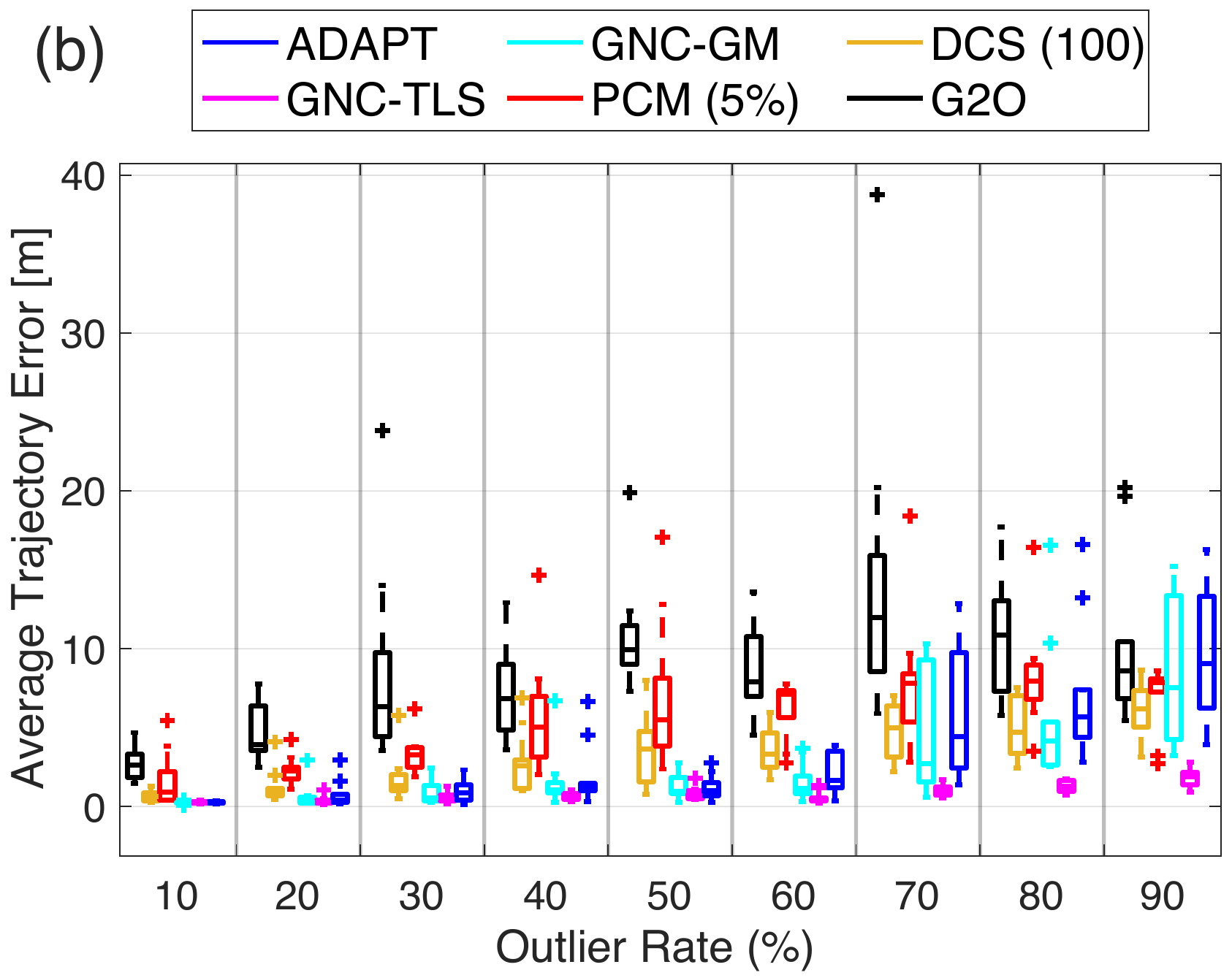} \\
			\end{minipage}
		& \hspace{-5mm}
			\begin{minipage}{\mpwthree}%
			\centering%
			\includegraphics[width=\columnwidth]{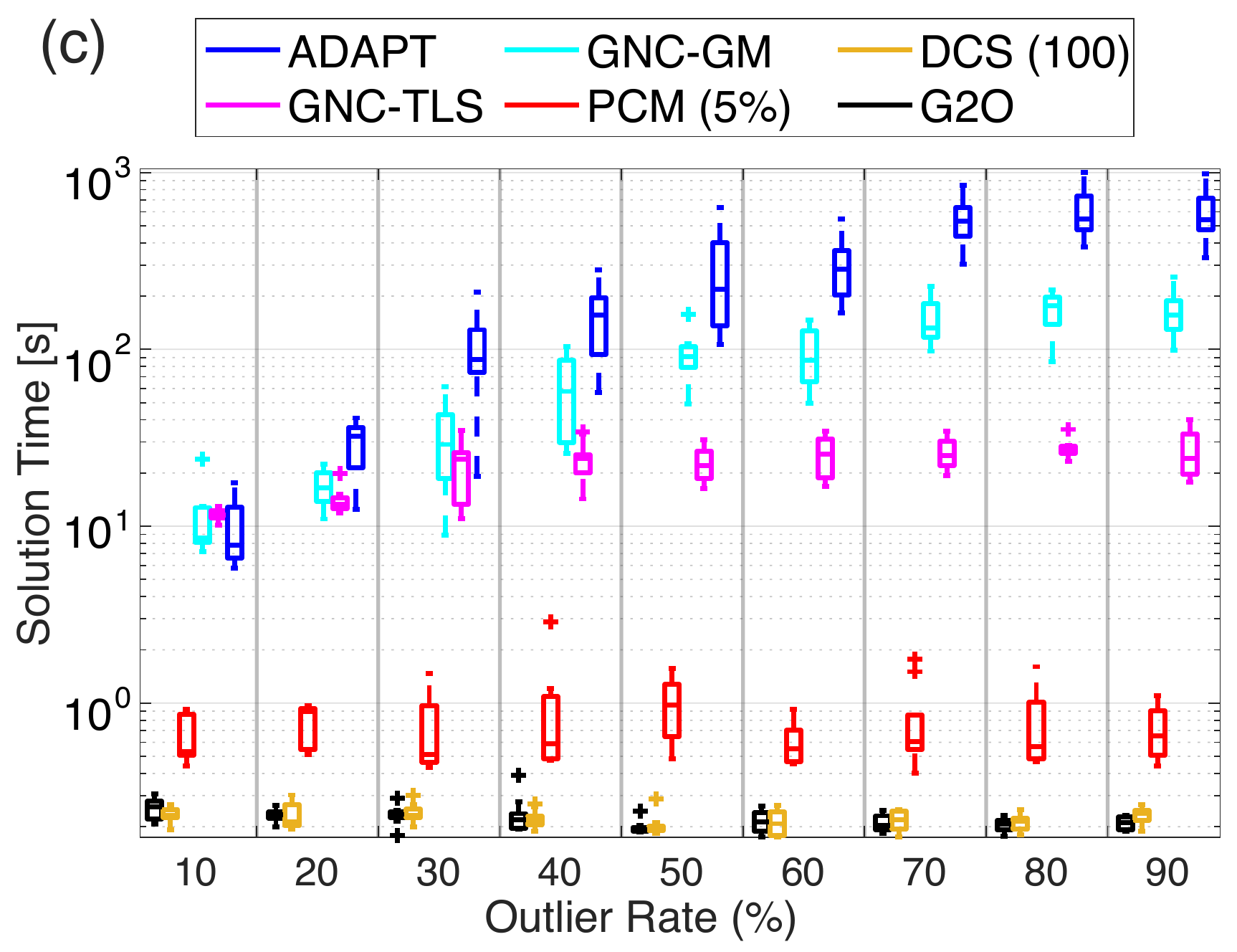} \\
			\end{minipage}
		\end{tabular}
	\end{minipage}
	\vspace{-3mm} 
	\caption{{\bf Pose Graph Optimization.} Performance of \GNCGM and \GNCTLS compared with state-of-the-art techniques for increasing outliers.  Average trajectory error for (a) \intel dataset and (b) \csail dataset; (c) solution time on the \csail dataset. Statistics are computed over 10 Monte Carlo runs.\label{fig:PGO}} 
	\vspace{-9mm} 
	\end{center}
\end{figure*}

\myParagraph{Point Cloud Registration Results} We use the \bunny dataset from the Stanford 3D Scanning Repository~\cite{Curless96siggraph}. We first scale the \bunny point cloud to be inside a unit cube, and then \finalize{at each Monte Carlo run we} apply a random rotation and translation to get a transformed copy of the \bunny. $N=100$ correspondences are randomly chosen, where we add zero-mean Gaussian noise with standard deviation $\sigma=0.01$ to the inliers, while corrupt the outliers with randomly generated points as in~\cite{Yang19rss-teaser}. We benchmark the performance of \GNCGM and \GNCTLS against (i) \ransac with \revise{1000} maximum iterations~\finalize{and $99\%$ confidence} using Horn's 3-point minimal solver~\cite{Horn87josa},\revise{~plus refinement using the maximum consensus set of inliers,} (ii) \adapt~\cite{Tzoumas19iros-outliers}, and (iii) \TEASER~\cite{Yang19rss-teaser}. 

Fig.~\ref{fig:pointCloudRegistration} reports the statistics for $60\%$-$95\%$ outlier rates (all methods work well below $60\%$). 
 Fig.~\ref{fig:pointCloudRegistration}(a)-(b) show the rotation and translation errors for increasing outliers. 
\revise{\ransac, \GNCGM, \GNCTLS, and \adapt all break at 90\% outliers and achieve similar estimation accuracy at outlier ratio below 90\%.} 
\TEASER is a specialized robust \emph{global} solver and outperforms all other techniques; 
 unfortunately, it currently does not scale to large problem instances\revise{~(>5min runtime)} and does not extend to other registration problems (\eg mesh registration); \revise{Fig.~\ref{fig:pointCloudRegistration}(c) plots the number of inner iterations 
  used by \ransac, \GNCGM, \GNCTLS, and \adapt (\TEASER has no outlier iterations). With respect to the outlier rates, the number of iterations grows very fast for \ransac, grows linearly for \adapt and is almost constant for \GNCGM and \GNCTLS. At $80\%$ outliers, the average runtimes for \ransac, \GNCGM, and \GNCTLS are $218$, $22$ and $23$ms, respectively, showing that \GNC can be much faster than \ransac.}
 For point cloud registration, \GNCGM is essentially the same as Fast Global Registration~\cite{Zhou16eccv-fastGlobalRegistration}, while below we show that the use of non-minimal solvers allows extending 
 \GNC to other spatial perception applications.

\myParagraph{Mesh Registration Results} We apply \GNCGM and \GNCTLS to register a point cloud to a mesh, 
 using the non-minimal solver~\cite{Briales17cvpr-registration}. 
 We use the ``car-2'' mesh model from the \PASCALplus dataset~\cite{Xiang2014WACV-PASCAL+}. \finalize{At each Monte Carlo run}, we generate a point cloud from the mesh by randomly sampling points lying on the vertices, edges and faces of the mesh model, and then apply a random transformation and add Gaussian noise with $\sigma=0.05$. We establish 40 point-to-point, 80 point-to-line and 80 point-to-plane correspondences, and create outliers by adding 
 incorrect point to point/line/plane 
correspondences (Fig.~\ref{fig:meshRegistration}{(d)}). 
We benchmark \GNCGM and \GNCTLS against (i) \ransac with 10,000 maximum iterations~\finalize{and $99\%$ confidence} using the 12-point minimal solver in~\cite{Khoshelham16jprs-ClosedformSolutionPlaneCorrespondences} and (ii) \adapt~\cite{Tzoumas19iros-outliers}. 

Fig.~\ref{fig:meshRegistration}{(a)-(c)} show the errors and iterations for each technique. \GNCGM, \GNCTLS, and \adapt are robust against 80\% outliers, while \ransac breaks at 50\% outliers. The number of iterations of \adapt grows linearly with the number of outliers, while \GNCGM's and \GNCTLS's iterations remain constant. 
Fig.~\ref{fig:meshRegistration}{(e)} shows a successful registration with \GNCTLS and 
Fig.~\ref{fig:meshRegistration}{(f)} shows an incorrect registration from \ransac, 
both obtained in a test with 70\% outliers. 

\subsection{Pose Graph Optimization}
\label{sec:PGO}

\myParagraph{Setup} \PGO is one of the most common estimation engines for SLAM~\cite{Cadena16tro-SLAMsurvey}. 
 \PGO estimates a set of poses $(\vt_i,\MR_i)$, $i=1,\ldots,n$ (typically sampled along the robot trajectory) 
from pairwise relative pose measurements $(\tldvt_{ij},\tldMR_{ij})$ (potentially corrupted with outliers). 
The residual error is the distance 
between the expected relative pose and the measured one: 
\bea
r(\{\MR_i,\vt_i\}) \!
= \!\sqrt{\kappa_{ij} \| \MR_j - \MR_i \tldMR_{ij} \|_F^2 + \tau_{ij} \| \vt_j - \vt_i - \MR_i \tldvt_{ij} \|^2_2},
\nonumber
\eea 
where $\kappa_{ij}$ and $\tau_{ij}$ are known parameters describing the measurement noise distribution, 
and $\|\cdot\|_F$ denotes the Frobenious norm. \sesync~\cite{Rosen18ijrr-sesync} 
 is a fast non-minimal solver for \PGO.

\myParagraph{PGO Results} 
We test \GNCGM and \GNCTLS on two standard benchmarking datasets: \intel and \csail, described in~\cite{Rosen18ijrr-sesync}. In each dataset, we preserve the odometry measurements, but~\finalize{at each Monte Carlo run} we spoil loop closures with random outliers. 
To create outliers, we sample random pairs of poses and add a random measurement between them.
We benchmark \GNCGM and \GNCTLS against (i) \gtwoo~\cite{Kuemmerle11icra}, 
(ii) \emph{dynamic covariance scaling} (\dcs)~\cite{Agarwal13icra}, 
(iii) \emph{pairwise consistent measurement set maximization} (\pcm)~\cite{Mangelson18icra}, 
and (iv) \adapt~\cite{Tzoumas19iros-outliers}.
\dcs and \pcm are fairly sensitive to the choice of parameters: 
  we tested parameters $\Phi =\{1,10,100\}$ for \dcs, and thresholds $\tau = \{5\%,10\%,15\%\}$ for \pcm, 
  and, for the sake of clarity, we only reported the choice of parameters ensuring the best performance.

Fig.~\ref{fig:PGO}(a) shows the average trajectory errors for the \intel dataset. 
\gtwoo is not a robust solver and performs poorly across the spectrum. 
\dcs and \pcm are specialized robust local solvers, 
but their errors gradually increase with the percentage 
of outliers. 
\GNCGM, \GNCTLS, and \adapt are insensitive to up to $40\%$ outliers and preserve an acceptable performance 
till $70-80\%$ of outliers; 
 \GNCTLS dominates the others. 
 Fig.~\ref{fig:PGO}(b) reports the results on the \csail dataset. Also in this case, 
 \GNCTLS dominates the other techniques and is robust to 90\% outliers. 
Fig.~\ref{fig:PGO}(c) reports the CPU times required by the techniques to produce a solution on \csail; 
in this case, all techniques are implemented in C++. 
\isExtended{While \adapt becomes impractical for problem with many outliers, 
\GNCTLS is fast and requires a time comparable to \dcs and \pcm, while affording improved robustness.}


\begin{figure*}[h]
	\begin{center}
	\vspace{-3mm}
	\begin{minipage}{\textwidth}
	\begin{tabular}{ccc}%
		\hspace{-5mm}
			\begin{minipage}{\mpwthree}%
			\centering%
			\includegraphics[width=\columnwidth]{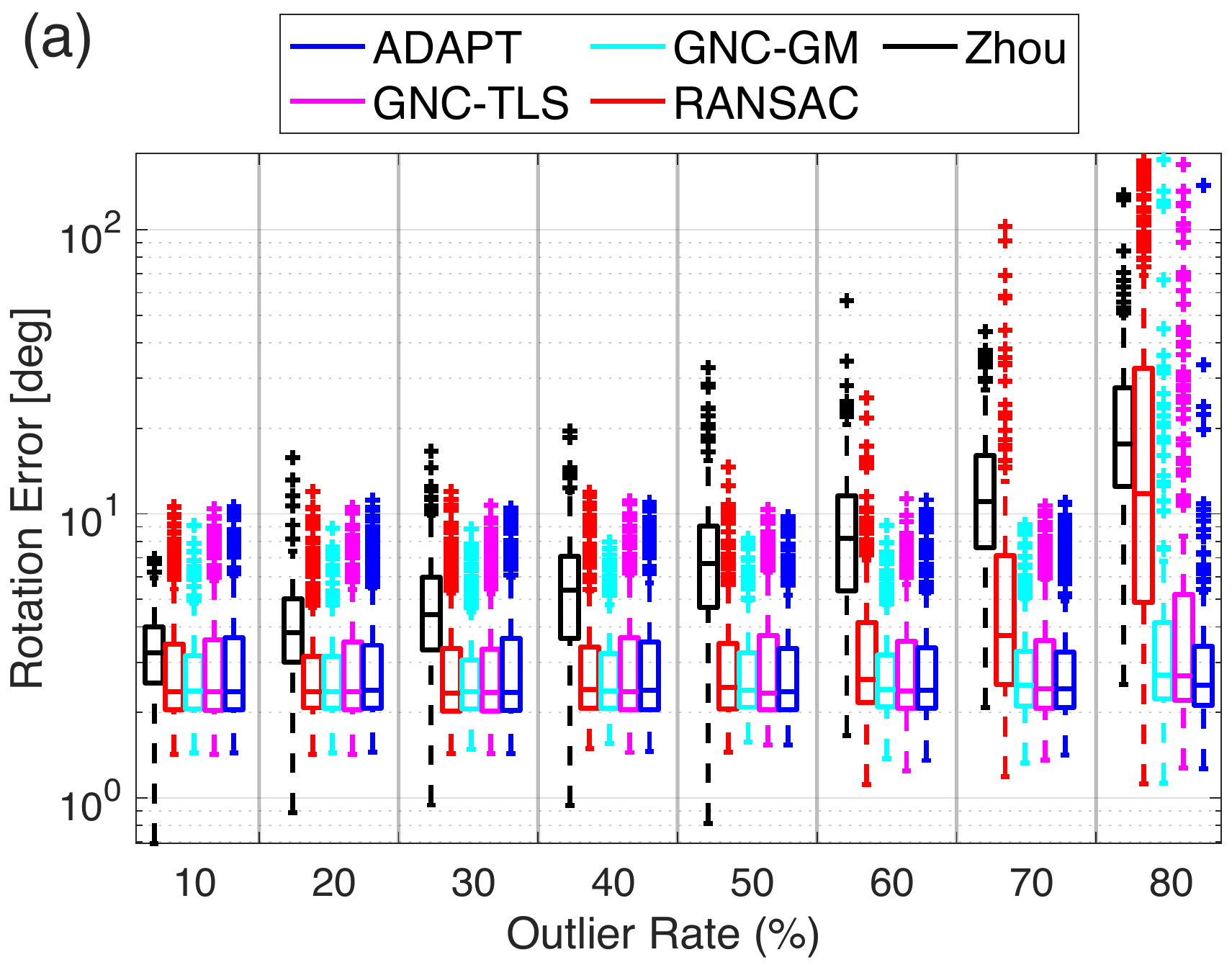} \\
			\end{minipage}
		& \myhspace
			\begin{minipage}{\mpwthree}%
			\centering%
			\includegraphics[width=\columnwidth]{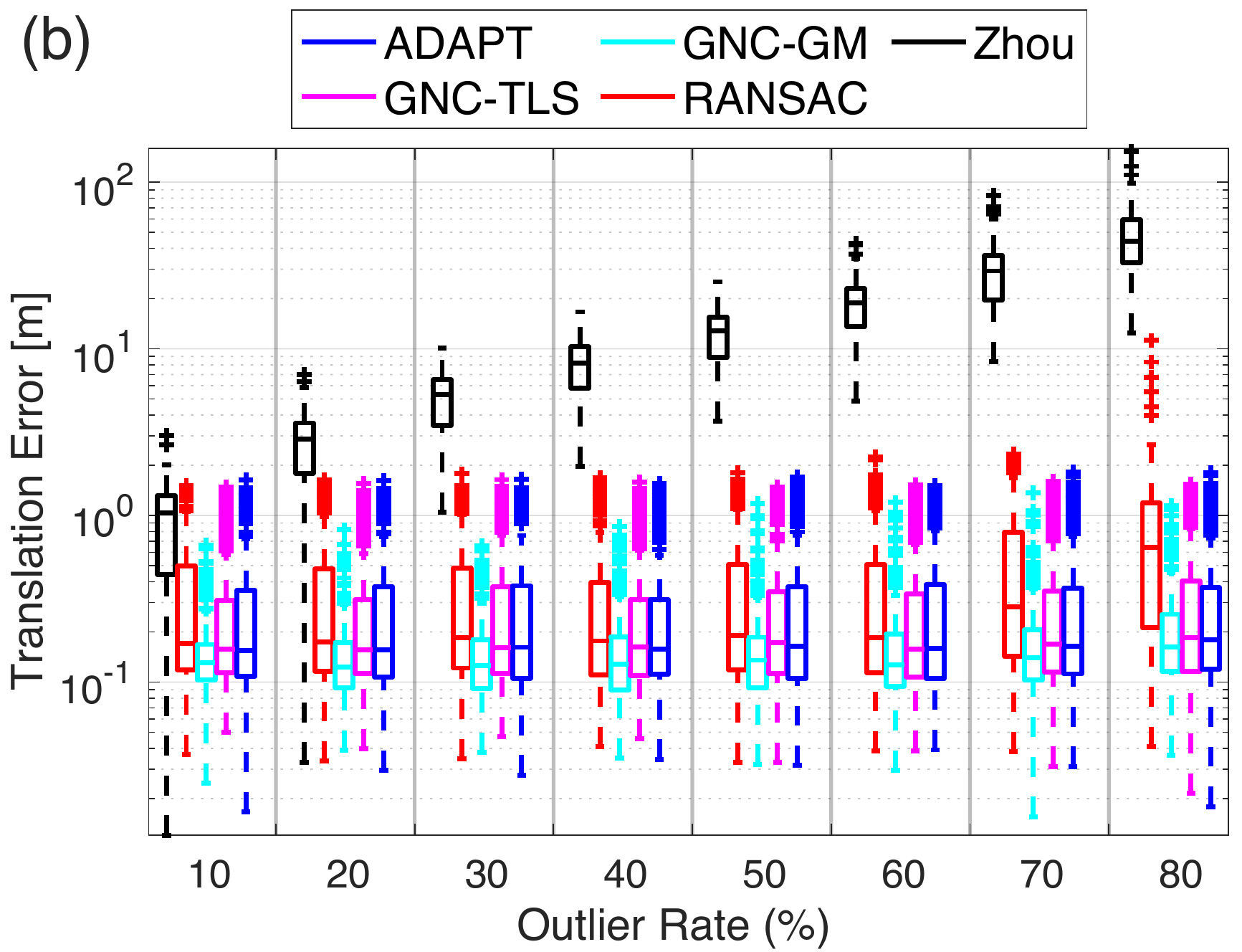} \\
			\end{minipage}
		& \hspace{-5mm}
			\begin{minipage}{\mpwthree}%
			\centering%
			\includegraphics[width=\columnwidth]{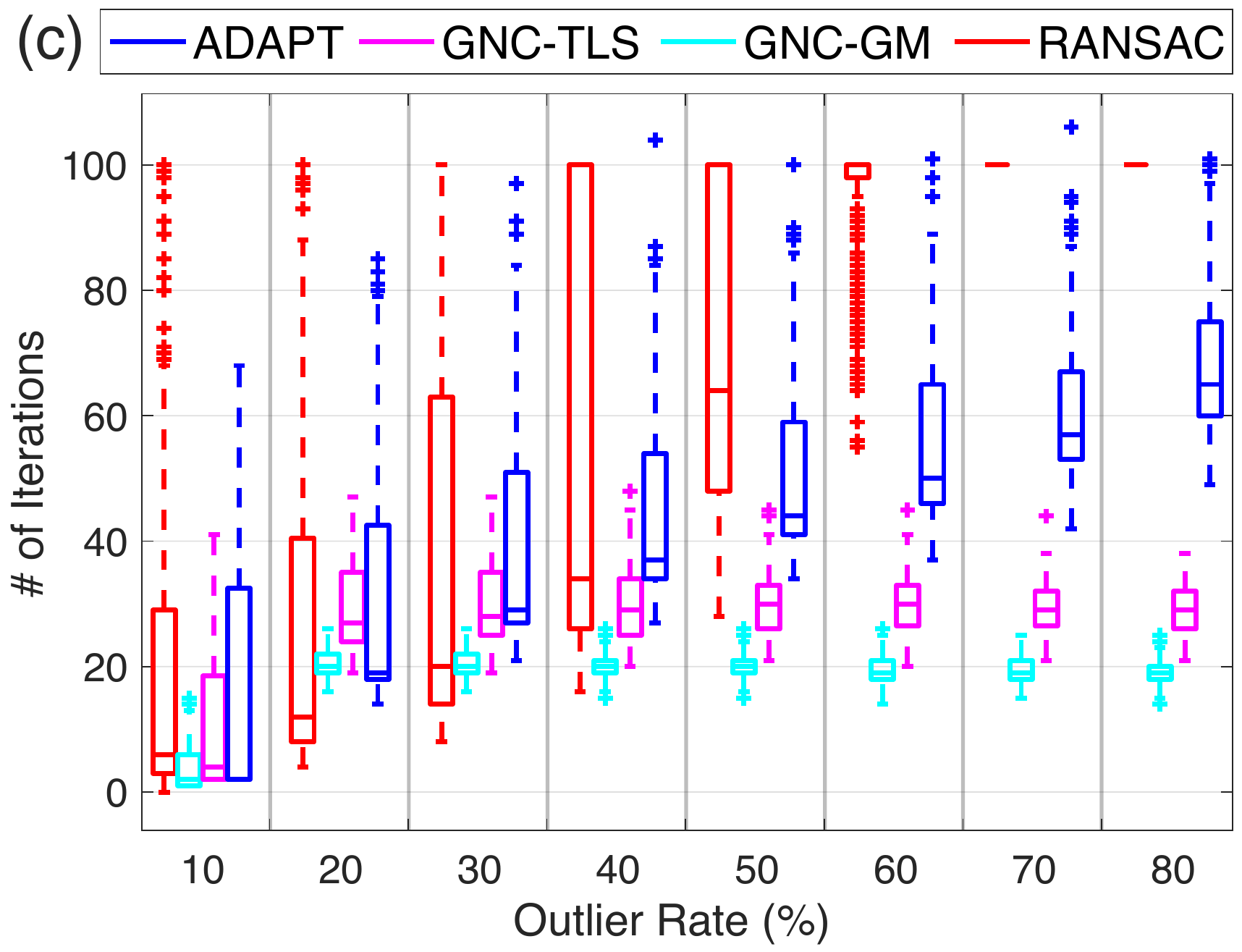} \\
			\end{minipage} \\ 

		\hspace{-4mm}
			\begin{minipage}{\mpwthree}%
			\centering%
			\includegraphics[width=\columnwidth]{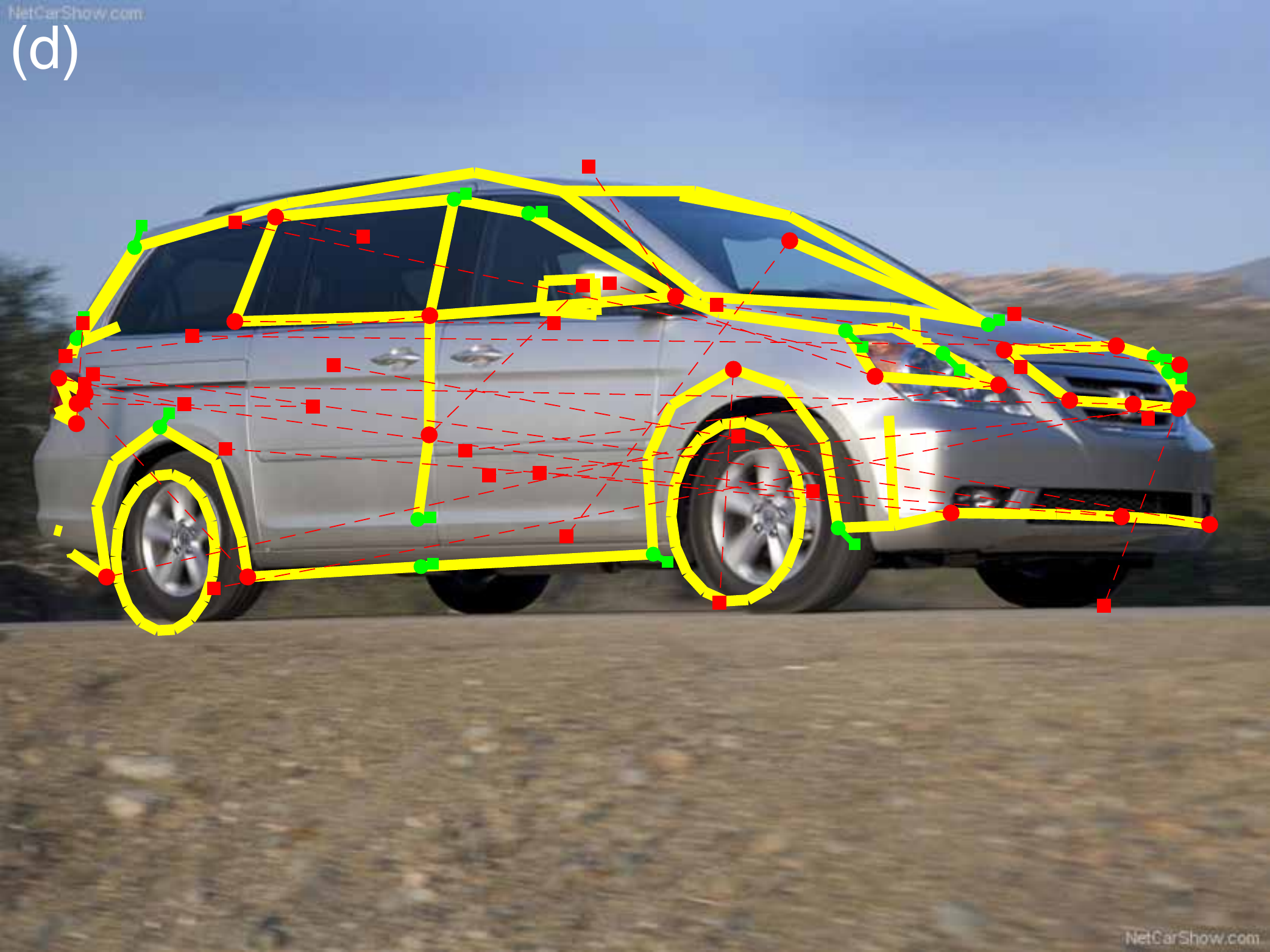} \\
			\end{minipage}
		& \myhspace
			\begin{minipage}{\mpwthree}%
			\centering%
			\includegraphics[width=\columnwidth]{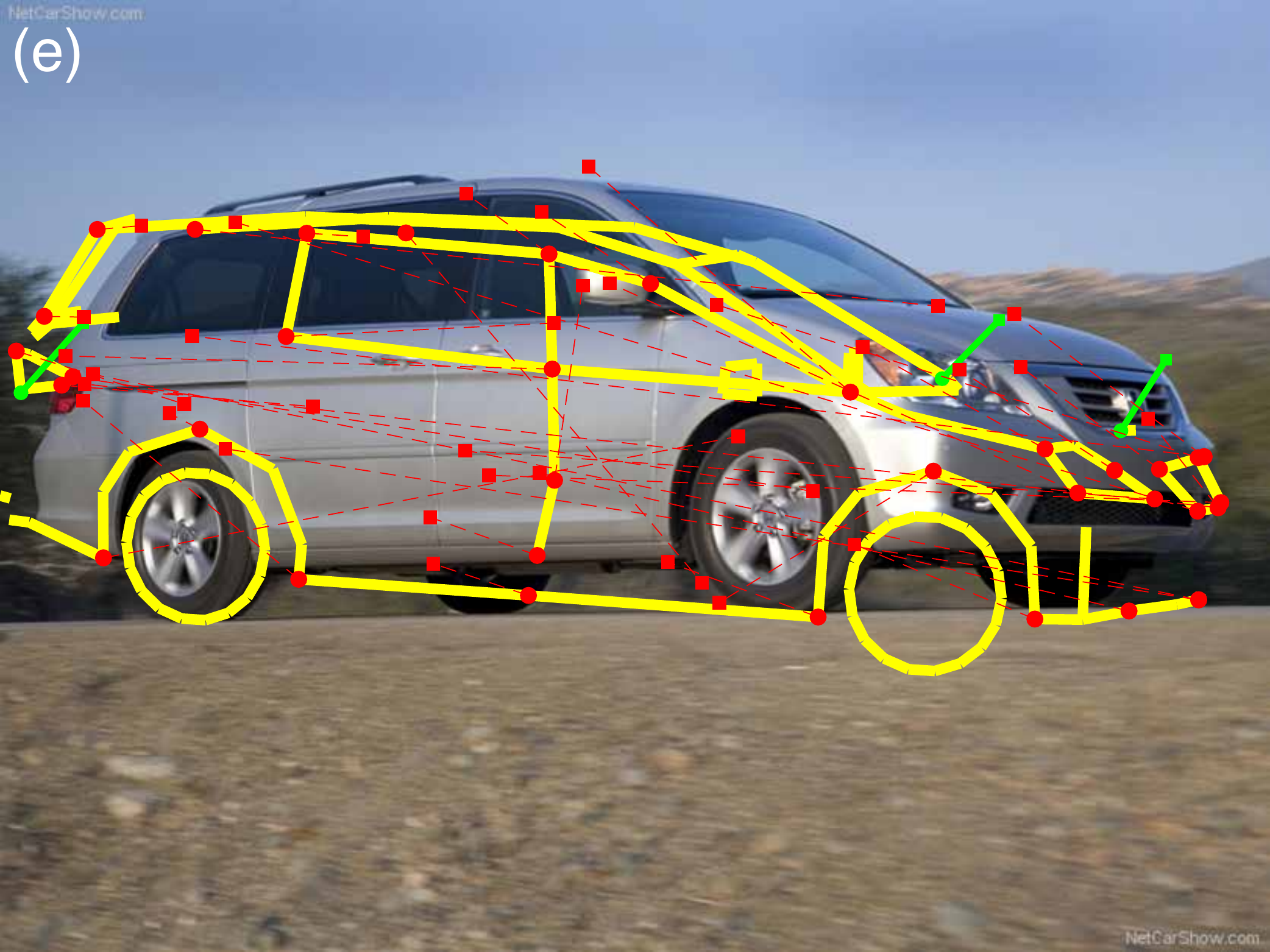} \\
			\end{minipage}
		& \hspace{-4.5mm}
			\begin{minipage}{\mpwthree}%
			\centering%
			\includegraphics[width=\columnwidth]{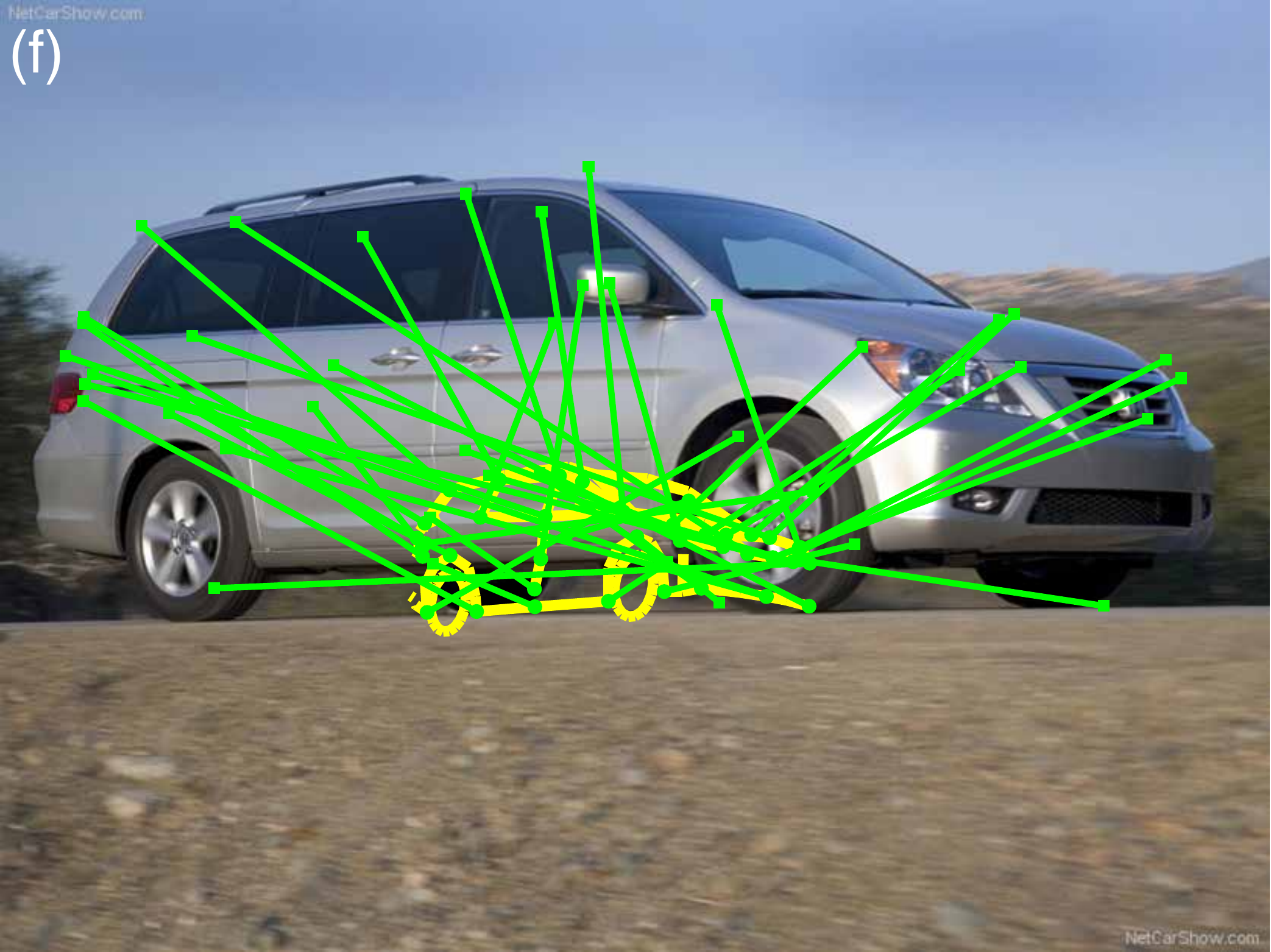} \\
			\end{minipage} \\ 

		\end{tabular}
	\end{minipage}
	\vspace{-3mm} 
	\caption{
{\bf Shape Alignment.} Performance of \GNCGM and \GNCTLS compared with state-of-the-art techniques on the \FGCar dataset~\cite{Lin14eccv-modelFitting} for increasing outliers.  
Quantitative results: (a) rotation error; (b) translation error; (c) number of iterations. 
Qualitative results with 70\% outliers: (d) successful pose estimation by \GNCGM; 
(e) failed pose estimation by \ransac; (f) failed pose estimation by Zhou's method~\cite{Zhou17pami-shapeEstimationConvex}. 
Yellow thick line: 3D skeleton of the car model; circle: 3D points on the car model; square: detected 2D features on the image; green: inlier correspondences; red: outlier correspondences. 
	\label{fig:shapeAlignment}}
	\vspace{-9mm} 
	\end{center}
\end{figure*}

\subsection{Shape Alignment}
\label{sec:SA} 

\myParagraph{Setup} 
In shape alignment, given 2D features $\vz_i \in \Real{2}, i=1,\ldots,N$, on a single image and 3D points $\MB_i \in \Real{3}, i=1,\ldots,N$,
of an object with putative correspondences $\vz_i \leftrightarrow \MB_i$ (potentially including outliers), the goal is to find the best scale $s >0$, rotation $\MR$, and translation $\vt$ of the object, 
 that projects the 3D shape to the 2D image under weak perspective projection. The residual function is $r(s,\MR,\vt) = \| \vz_i - s\Pi\MR \MB_i - \vt \|$, where $\Pi \in \Real{2 \times 3}$ is the weak perspective projection matrix (equal to the first two rows of a $3\!\times\!3$ identity matrix). 
 Note that $\vt$ is a 2D translation, but under weak perspective projection one can extrapolate a 3D translation 
 (\ie recover the distance of the camera to the object) using the scale $s$. 

\myParagraph{A Non-minimal Solver for Shape Alignment} 
 The literature is missing a global solver for shape alignment, even in the outlier-free case. Therefore, we start by proposing a 
 non-minimal solver for (weighted) outlier-free shape alignment: 
\begin{equation} \label{eq:shapeAlignment}
\min_{s>0, \MR \in \SOthree, \vt \in \Real{2}} \textstyle{\sumAllPointsi} w_i \| \vz_i - s\Pi\MR \MB_i - \vt \|^2, 
\end{equation}
where $w_i$ are constant weights.
The next proposition states that the global minimizer of problem~\eqref{eq:shapeAlignment} can be obtained by solving a quaternion-based unconstrained optimization.

\begin{proposition}[Quaternion-based Shape Alignment] \label{prop:quaternionRotationShapeAlignment}
Define the non-unit quaternion $\vv \doteq \sqrt{s}\vq \in \Real{4}$, where $\vq$ is the unit-quaternion corresponding to $\MR$ and $s$ is the unknown scale.
 Moreover, if $\vv = [v_1, v_2, v_3, v_4]\tran$, define $[\vv]_2 \doteq [v_1^2,v_2^2,v_3^2,v_4^2,v_1v_2,v_1v_3,v_1v_4,v_2v_3,v_2v_4,v_3v_4]\tran \in \Real{10}$ as the vector of degree-2 monomials in $\vv$.
  Then the globally optimal solutions ($s^\star, \MR^\star, \vt^\star$) of problem~\eqref{eq:shapeAlignment} can be obtained from the  solution $\vv^\star$ of the following optimization:
\begin{equation} \label{eq:quaternionRotationShapeAlignment}
\min_{\vv \in \Real{4}} f(\vv) \doteq [\vv]_2\tran \MQ [\vv]_2 - 2 \vg\tran [\vv]_2 + h, 
\end{equation}
where $\MQ \in \Real{10\times10}$, $\vg \in\Real{10}$, and $h>0$ are known quantities, whose expression is given in the supplementary material~\cite{Yang20tr-GNC}.
\end{proposition}

Problem~\eqref{eq:quaternionRotationShapeAlignment} requires minimizing  a degree-4 polynomial $f(\vv)$ in 4 variables; 
to this end, we apply SOS relaxation and relax~\eqref{eq:quaternionRotationShapeAlignment} to the following SOS optimization:
\bea \label{eq:sosqShapeAlign}
\min_{ \vv \in \Real{4}, \gamma \in \Real{} }  & -\gamma \quad , \quad \text{s.t.}\quad f(\vv) - \gamma \text{ is SOS},
\eea
which can be readily converted to an SDP and solved with certifiable optimality~\cite{Blekherman12Book-sdpandConvexAlgebraicGeometry}. We use the \gPoly~\cite{henrion2009optimmethodsoftw-gloptipoly3} package in Matlab to solve problem~\eqref{eq:sosqShapeAlign} and found that empirically the relaxation is always exact. 
 Solving the SDP takes about 80 ms on a desktop computer.


\myParagraph{Shape Alignment Results} 
We test the performance of \GNCGM and \GNCTLS, together with our SOS solver on the \FGCar dataset~\cite{Lin14eccv-modelFitting}, where we use the ground-truth 3D shape model as $\MB$ and the ground-truth 2D landmarks as $\vz$. 
To generate outliers~\finalize{for each image}, we set random incorrect correspondences between 3D points and 2D features. 
We benchmark \GNCGM and \GNCTLS against (i) Zhou's method~\cite{Zhou17pami-shapeEstimationConvex}, (ii) \ransac with 100 maximum iterations~\finalize{and $99\%$ confidence} using a 4-point minimal solver (we use our SOS solver as minimal solver), and (iii) \adapt~\cite{Tzoumas19iros-outliers}. 

Fig~\ref{fig:shapeAlignment}(a)-(c) show translation errors, rotation errors, and number of iterations for all compared techniques. 
 Statistics are computed over all 600 images in the \FGCar dataset. The performance of Zhou's method degrades quickly with increasing outliers.
  \ransac breaks at 60\% outliers. 
  \GNCGM, \GNCTLS, and \adapt are robust against 70\% outliers, while \GNCGM and \GNCTLS require a roughly constant number of iterations. 
  Qualitative results for \GNCGM, \ransac, and Zhou's approach are given in Fig.~\ref{fig:shapeAlignment}(d)-(f), respectively.

\section{Conclusions}
\label{sec:conclusions}

We proposed a general-purpose approach for robust estimation that leverages modern non-minimal solvers. 
The approach allows extending the applicability of Black-Rangarajan duality and Graduated Non-convexity (\GNC) 
to several spatial perception problems, ranging from mesh registration and shape alignment to pose graph optimization.
We believe the proposed approach can be a valid replacement for \ransac. 
While \ransac requires a minimal solver, our \GNC approach requires a \emph{non-minimal} solver. 
 Our approach is deterministic, resilient to a large number of outliers, 
 and significantly faster than specialized solvers. 
 As a further contribution, we presented a non-minimal solver for shape alignment.
  \revise{Future work includes investigating \emph{a priori} and \emph{a posterior} conditions that 
  guarantee convergence of \GNC to globally optimal solutions.}

\clearpage

\begin{center}
\large{{\bf Supplementary Material}}
\end{center}

\renewcommand{\theequation}{A\arabic{equation}}

\section{Proof of Proposition~\ref{prop:GNCGM}}
The outlier process~\eqref{eq:outlierProcessGNCGM} is derived by following the Black-Rangarajan procedure in Fig.~10 of~\cite{Black96ijcv-unification}. Therefore, we here prove the weight update rule in eq.~\eqref{eq:dualWeightUpdateGNCGM}. To this end, we derive the gradient $g_i$ of the objective function with outlier process $\Phi_{\rho_{\mu}}(w_i) = \mu\barcsq(\sqrt{w_i} - 1)^2$ in eq.~\eqref{eq:weightUpdate} with respect to $w_i$:
\bea\label{eq:aux_1}
g_i = \hatr_i^2 + \mu \barcsq \left( 1 - \frac{1}{\sqrt{w_i}} \right).
\eea
From eq.~\eqref{eq:aux_1} we observe that if $w_i \rightarrow 0$, then $g_i \rightarrow -\infty$; and if $w_i = 1$, then $g_i = \hatr_i^2 \geq 0$. These facts, combined with the monotonicity of $g_i$ \wrt $w_i$, ensure that there exists a unique $w_i^\star$ such that the gradient $g_i$ vanishes:
\bea \label{eq:dualUpdateGM}
w_i^\star = \left( \frac{\mu \barcsq}{\hatr_i^2 + \mu\barcsq} \right)^2.
\eea
This vanishing point is the global minimizer of~\eqref{eq:weightUpdate}.
\section{Proof of Proposition~\ref{prop:GNCTLS}}
The outlier process~\eqref{eq:outlierProcessGNCTLS} is derived by following the Black-Rangarajan procedure in Fig.~10 of~\cite{Black96ijcv-unification}. Therefore, we here prove the weight update rule in eq.~\eqref{eq:dualWeightUpdateGNCTLS}. To this end, we derive the gradient $g_i$ of the objective function with outlier process $\Phi_{\rho_{\mu}}(w_i) = \frac{\mu (1-w_i)}{ \mu + w_i}$ in eq.~\eqref{eq:weightUpdate} with respect to $w_i$:
\bea\label{eq:aux_2}
 g_i = \hatr_i^2 - \frac{\mu(\mu+1)}{(\mu + w_i)^2}\barcsq.
\eea
From eq.~\eqref{eq:aux_2} we observe that if $w_i = 0$, then $g_i = \hatr_i^2 - \frac{\mu+1}{\mu} \barcsq$; and if $w_i = 1$, then $g_i = \hatr_i^2 - \frac{\mu}{\mu+1} \barcsq$. Therefore, the global minimizer $w_i^\star$ can be obtained by setting the gradient $g_i$ to zero, leading to:
\bea \label{eq:dualUpdateTLS}
w_i^\star = \begin{cases}
0 & \text{ if }\hatr_i^2 \in \left[ \frac{\mu+1}{\mu}\barcsq, +\infty \right] \\
\frac{\barc}{\hatr_i}\sqrt{\mu(\mu+1)} - \mu & \text{ if } \hatr_i^2 \in \left[ \frac{\mu}{\mu+1}\barcsq ,\frac{\mu+1}{\mu}\barcsq \right] \\
1 & \text{ if } \hatr_i^2 \in \left[ 0,\frac{\mu}{\mu+1}\barcsq \right]
\end{cases}.
\eea


\section{Proof of Proposition~\ref{prop:quaternionRotationShapeAlignment}}
In the main document, we claim that the optimal solutions $(s^\star, \MR^\star, \vt^\star)$ for the shape alignment problem
\begin{equation} \label{eq:shapeAlignmentSupp}
\min_{s>0, \MR \in \SOthree, \vt \in \Real{2}} \textstyle{\sumAllPointsi} w_i \| \vz_i - s\Pi\MR \MB_i - \vt \|^2 
\end{equation}
can be obtained from the optimal solution $\vv^\star$ of the quaternion-based unconstrained optimization:
\begin{equation} \label{eq:quaternionShapeAlignmentSupp}
\min_{\vv \in \Real{4}} f(\vv) \doteq [\vv]_2\tran \MQ [\vv]_2 - 2 \vg\tran [\vv]_2 + h. 
\end{equation}
Here we prove this proposition, and provide a formula to compute $\MQ \in \Real{10 \times 10}$, $\vg \in \Real{10}$ and $h$, as well as construct $(s^\star, \MR^\star, \vt^\star)$ from $\vv^\star$.  Particularly, we complete the proof in two steps: (i) we marginalize out the translation $\vt$ and convert problem~\eqref{eq:shapeAlignmentSupp} into an equivalent translation-free problem; (ii) we reparametrize the rotation matrix using unit-quaternion and then arrive at the unconstrained optimization~in~\eqref{eq:quaternionShapeAlignmentSupp}.  The two steps follow in more detail below:

{\bf (i) Translation-free Shape Alignment.} To this end, we develop the objective function of~\eqref{eq:shapeAlignmentSupp}:
\bea
& \sumAllPointsi w_i \| \vz_i - s\Pi \MR \MB_i - \vt \|^2  \nonumber \\
& = \sumAllPointsi w_i \vt\tran \vt - 2w_i (\vz_i\tran - s\MB_i\tran \MR\tran \Pi\tran) \vt \nonumber \\
&+  \sumAllPointsi \| \vz_i - s\Pi \MR \MB_i\|^2,
\eea
whose partial derivative \wrt $\vt$ is:
\bea \label{eq:derivativet}
2\sumAllPointsi w_i \vt - 2\left(\sumAllPointsi  w_i \vz_i - s \Pi \MR  \sumAllPointsi w_i \MB_i \right).
\eea
By setting the derivative~\eqref{eq:derivativet} to zero, we derive $\vt^\star$ in closed form as:
\bea \label{eq:closedFormt}
\vt^\star = \bar{\vz}_w - s^\star \Pi \MR^\star \bar{\MB}_w,
\eea
where
\bea
\bar{\vz}_w = \frac{\sumAllPointsi w_i \vz_i}{ \sumAllPointsi w_i },\quad \bar{\MB}_w = \frac{\sumAllPointsi w_i \MB_i}{ \sumAllPointsi w_i }.
\eea
By inserting eq.~\eqref{eq:closedFormt} back to the original problem~\eqref{eq:shapeAlignmentSupp}, we obtain an optimization only involving the scale and rotation:
\bea \label{eq:rotationOnlyShapeAlignmentSupp}
\min_{s>0, \MR \in \SOthree} \sumAllPointsi \| \tldvz_i - s\Pi\MR \tldMB_i \|^2,
\eea
where:
\bea
\tldvz_i = \sqrt{w_i}(\vz_i - \barvz_{w}),\quad \tldMB_i = \sqrt{w_i} (\MB_i - \barMB_{w}).
\eea

{\bf (ii) Quaternion-based Translation-free Problem.} We now show that the translation-free shape alignment problem~\eqref{eq:rotationOnlyShapeAlignmentSupp} can be converted to the quaternion-based unconstrained shape alignment problem~\eqref{eq:quaternionShapeAlignmentSupp}. To this end, we denote $\vq \in \Real{4}, \| \vq \| =1$ as the unit-quaternion corresponding to the rotation matrix $\MR$.  It can be verified by inspection that $\MR$ is a linear function of $[\vq]_2 \in \Real{10}$, the vector of all degree-2 monomials of $\vq$:
\bea \label{eq:Rfromq}
\MR = \calR(\vq) \doteq \text{mat} ( \MA [\vq]_2 ),
\eea
where $\MA \in \Real{9 \times 10}$ is a constant matrix, and $\text{mat}(\cdot)$ converts a vector to a matrix with proper dimension (in this case a $3 \times 3$ matrix). The expression of the matrix $\MA$ is as follows:
\bea
\hspace{-4mm} \MA = \left[ \begin{array}{cccccccccc}
1 & -1 & -1 & 1 & 0 & 0 & 0 & 0 & 0 & 0 \\
0 & 0 & 0 & 0 & 2 & 0 & 0 & 0 & 0 & 2 \\
0 & 0 & 0 & 0 & 0 & 2 & 0 & 0 & -2 & 0 \\
0 & 0 & 0 & 0 & 2 & 0 & 0 & 0 & 0 & -2 \\
-1 & 1 & -1 & 1 & 0 & 0 & 0 & 0 & 0 & 0 \\
0 & 0 & 0 & 0 & 0 & 0 & 2 & 2 & 0 & 0 \\
0 & 0 & 0 & 0 & 0 & 2 & 0 & 0 & 2 & 0 \\
0 & 0 & 0 & 0 & 0 & 0 & -2 & 2 & 0 & 0 \\
-1 & -1 & 1 & 1 & 0 & 0 & 0 & 0 & 0 & 0
\end{array} \right].
\eea
Using the notation in~\eqref{eq:Rfromq} and developing the squares in the objective function of problem~\eqref{eq:rotationOnlyShapeAlignmentSupp}\revise{~(denoting $\vr = \vectorize{\MR}$)}:
\bea
\hspace*{6mm}& \displaystyle \sumAllPointsi \| \tldvz_i - s\Pi\MR\tldMB_i \|^2 \nonumber \\
& = \displaystyle \sumAllPointsi  \| \tldvz_i \|^2 - 2 s\; \tldvz_i\tran \Pi \MR \tldMB_i + s^2 \tldMB_i\tran \MR\tran \Pi\tran \Pi \MR \tldMB_i \nonumber \\
& \hspace{-8mm} = \displaystyle \sumAllPointsi \| \tldvz_i \|^2 - 2s \; \trace{\tldMB_i\vz_i\tran\Pi \MR} + s^2 \trace{\tldMB_i\tldMB_i\tran \MR\tran \Pi\tran \Pi \MR} \nonumber \\
& \hspace{-10mm} = \displaystyle \sumAllPointsi \|\tldvz_i\|^2 - 2s\; \vectorize{\Pi\tran \tldvz_i \tldMB_i\tran}\tran \vr + s^2 \vr\tran (\tldMB_i\tldMB_i\tran \kron \Pi\tran\Pi) \vr \nonumber \\
& = \displaystyle \sumAllPointsi \|\tldvz_i\|^2 - 2s \; \vf_i\tran \MA [\vq]_2 + s^2 [\vq]_2\tran \MA\tran \MF_i \MA [\vq]_2 \nonumber \\
& = \displaystyle \sumAllPointsi \| \vz_i \|^2 - 2s \; \vg_i\tran [\vq]_2 + s^2 [\vq]_2\tran \MQ_i [\vq]_2 \nonumber \\
& = s^2 [\vq]_2\tran \MQ [\vq]_2 - 2s\; \vg\tran [\vq]_2 + h, \label{eq:developedObjFunc}
\eea
where the $\MQ$, $\vg$, and $h$ can be computed by:
\bea \label{eq:Qgh}
\begin{cases}
\MQ = \MA\tran \left( \sumAllPointsi  \tldMB_i \tldMB_i\tran \kron \Pi\tran \Pi \right) \MA; \\
\vg = \MA\tran \left( \sumAllPointsi \vectorize{\Pi\tran \tldvz_i \tldMB_i\tran} \right); \\
h = \sumAllPointsi \| \tldvz_i \|^2.
\end{cases}
\eea
Now, since $s>0$, we define $\vv \doteq \sqrt{s} \vq$ and it is straightforward to verify that $[\vv]_2 = s[\vq]_2$, and therefore the objective function in~\eqref{eq:developedObjFunc} is equal to the objective function of problem~\eqref{eq:quaternionShapeAlignmentSupp}. In addition, because $\vv \doteq \sqrt{s} \vq$, there is no constraint on $\vv$, because the unit-norm constraint of the quaternion $\vq$ disappears. 

After we solve problem~\eqref{eq:quaternionShapeAlignmentSupp}, we can recover the optimal solution $(s^\star, \MR^\star, \vt^\star)$ of the original constrained optimization~\eqref{eq:shapeAlignmentSupp} from the optimal solution $\vv^\star$ of problem~\eqref{eq:quaternionShapeAlignmentSupp} using the following formula:
\bea \label{eq:recoversRt}
\begin{cases}
s^\star = \| \vv^\star \|^2; \\
\MR^\star = \calR \left( \frac{\vv^\star}{\| \vv^\star \|} \right); \\
\vt^\star = \barvz_{w} - s^\star \Pi \MR^\star \barMB_{w}.
\end{cases}
\eea 
In summary, to solve the outlier-free (weighted) shape alignment problem~\eqref{eq:shapeAlignmentSupp}, we first calculate $\MQ$, $\vg$, $h$ using eq.~\eqref{eq:Qgh}, solve the unconstrained optimization~\eqref{eq:quaternionShapeAlignmentSupp} (using SOS relaxation as discussed in the main document), and then recover the optimal scale, rotation, translation using eq.~\eqref{eq:recoversRt}.

\bibliographystyle{IEEEtran}
\bibliography{refs,myRefs}
\end{document}